\documentclass[]{style}

\usepackage[toc,page,header]{appendix}
\usepackage{amsthm}

\usepackage{natbib}

\usepackage{CJKutf8}

\usepackage{xargs}  

\usepackage{todonotes}  

\usepackage{multirow}

\usepackage{cleveref}

\usepackage{amsmath}
\usepackage{dsfont}

\usepackage{svg}

\usepackage{mathrsfs}
\usepackage{adjustbox}
\usepackage{multirow}
\usepackage{multicol}
\usepackage{tcolorbox}
\usepackage{changepage}
\usepackage{enumitem}
\usepackage{graphicx}
\usepackage{amssymb}
\usepackage{xcolor}
\usepackage{float}
\usepackage{multirow}
\usepackage{threeparttable}
\usepackage{graphicx}
\usepackage{algorithm}
\usepackage{algpseudocode}
\usepackage{wrapfig}
\usepackage{epsfig}
\usepackage{overpic}
\usepackage{diagbox}
\usepackage{fontawesome}
\usepackage{bbding}
\usepackage{wasysym}
\usepackage{utfsym}
\usepackage{mathtools}
\usepackage[table]{xcolor}  
\usepackage{colortbl}       
\usepackage{tabularx} 
\usepackage{makecell} 
\usepackage[dvipsnames]{xcolor}

\newcolumntype{g}{>{\columncolor{gray!10}}c} 

\definecolor{catgray}{gray}{0.9}
\definecolor{skyblue}{rgb}{0.53,0.81,0.92} 

\colorlet{skyblue!30}{skyblue!30!white} 

\definecolor{customblue}{RGB}{70,130,180}  

\newtcolorbox{evolbox}[2][]{%
  enhanced,
  colframe=customblue,
  colback=white,
  coltitle=white,
  rounded corners,
  boxrule=1pt,
  titlerule=0pt,
  toptitle=1mm,
  bottomtitle=1mm,
  fonttitle=\bfseries,
  width=#2\textwidth, 
  #1
}

\usepackage{url}

\PassOptionsToPackage{table,xcdraw}{xcolor}
\usepackage{placeins}
\usepackage{pifont}

\definecolor{RowBlue}{HTML}{E9F2FB}
\definecolor{RowRed}{HTML}{F9EAEA}
\definecolor{Top1}{HTML}{50DB4B} 
\definecolor{Top2}{HTML}{A5FFA2} 
\definecolor{Top3}{HTML}{D9FFD9} 
\definecolor{Sub1}{HTML}{EAB8B8}
\definecolor{Sub2}{HTML}{E4E4E4}

\renewcommand{\emph}[1]{\textit{#1}}

\input{ursfs.fd}
\DeclareFontShape{U}{rsfs}{m}{n}{%
  <-7> rsfs5 <7-8> rsfs7 <8-> rsfs10}{}
\input{uwasy.fd}
\DeclareFontShape{U}{wasy}{m}{n}{%
  <-6> wasy5 <6-7> wasy6 <7-8> wasy7 <8-9> wasy8
  <9-10> wasy9 <10-> wasy10}{}

\usepackage{mdframed}

\mdfdefinestyle{propbox}{
    backgroundcolor=black!5,  
    linecolor=black,          
    linewidth=0.5pt,          
    innerleftmargin=8pt,      
    innerrightmargin=8pt,     
    innertopmargin=8pt,       
    innerbottommargin=8pt,    
    skipabove=10pt,           
    skipbelow=10pt            
}

\newmdtheoremenv[style=propbox]{proposition}{Proposition}
\newmdtheoremenv[style=propbox]{theorem}{Theorem}
\newmdtheoremenv[style=propbox]{lemma}{Lemma}

\usepackage[most]{tcolorbox}

\tcbuselibrary{theorems}

\newcounter{mydefinition}
\newtcolorbox[use counter=mydefinition]{definitionbox}[1][]{
  colback=blue!1!white,   
  colframe=myblue, 
  fonttitle=\bfseries,
  coltitle=black,
  title=Conclusion~\themydefinition, 
  boxrule=0.6pt,
  #1
}

\newtcolorbox[]{conclusionbox}[1][]{
  colback=green!1!white,
  colframe=mygreen,
  fonttitle=\bfseries,
  coltitle=black,
  title=Observation, 
  boxrule=0.6pt,
  #1
}

\title{STILL: Selecting Tokens for Intra-Layer Hybrid Attention to Linearize LLMs}
\headtitle{STILL: Selecting Tokens for Intra-Layer Hybrid Attention to Linearize LLMs}

\author[1,2,*]{Weikang Meng}
\author[1,*]{Liangyu Huo}
\author[3]{Yadan Luo}
\author[1]{Jiawen Guan}
\author[4]{Jingyi Zhang}
\author[2,\text{$\dagger$}]{\par{Yingjian Li}}
\author[1,\text{$\dagger$}]{Zheng Zhang}

\affiliation[1]{SMULL Group, Harbin Institute of Technology, Shenzhen}
\affiliation[2]{{Pengcheng Laboratory}\par}
\affiliation[3]{UQMM Lab, University of Queensland}
\affiliation[4]{Relty (Shanghai) Technology Co., Ltd.}

\contribution[*]{Equal contribution}
\contribution[\text{$\dagger$}]{Corresponding author}

\preprint{Preprint version.}
\abstract{
Linearizing pretrained large language models (LLMs) primarily relies on intra-layer hybrid attention mechanisms to alleviate the quadratic complexity of standard softmax attention.
Existing methods perform token routing based on sliding-window partitions, resulting in position-based selection and fails to capture token-specific global importance.
Meanwhile, linear attention further suffers from distribution shift caused by learnable feature maps that distort pretrained feature magnitudes.
Motivated by these limitations, we propose STILL, an intra-layer hybrid linearization framework for efficiently linearizing LLMs.
STILL introduces a Self-Saliency Score with strong local–global consistency, enabling accurate token selection using sliding-window computation, and retains salient tokens for sparse softmax attention while summarizing the remaining context via linear attention.
To preserve pretrained representations, we design a Norm-Preserved Feature Map (NP-Map) that decouples feature direction from magnitude and reinjects pretrained norms.
We further adopt a unified training–inference architecture with chunk-wise parallelization and delayed selection to improve hardware efficiency.
Experiments show that STILL matches or surpasses the original pretrained model on commonsense and general reasoning tasks, and achieves up to a 86.2\% relative improvement over prior linearized attention methods on long-context benchmarks. Code is available at \href{https://github.com/ZacharyMeng/STILL}{this URL}.
}

\correspondence{\email{zacharymengwk@gmail.com}, \email{darrenzz219@gmail.com}, \email{yingjian1122@gmail.com}}

\begin{document}

\maketitle

\vspace{-3ex}
\section{Introduction}
Transformer-based Large Language Models (LLMs) \citep{gpt3,llama} have achieved remarkable performance across a wide range of language tasks, largely owing to self-attention’s \citep{attention} ability to model long-range dependencies. 
However, vanilla self-attention requires computing pairwise interactions between all tokens, resulting in \textit{quadratic} complexity in both time and memory with respect to sequence length. This cost becomes a primary bottleneck as contemporary models push toward substantially longer contexts.
In response, linear attention \citep{linear_attention,gla,deltanet} has emerged as a promising alternative, enabling a better efficiency-performance balance and allowing LLMs to be scaled to much longer sequences \citep{minimax,kimilinear,qwen3} with higher speed and lower memory. 

\textcolor{black}{However, building these models from scratch remains computationally exorbitant. Recent efforts, such as MiniMax-Text-01 \citep{minimax} and Kimi-linear \citep{kimilinear}, attempt to offset the performance drop of pure linear attention by training \textit{\textbf{inter-layer} hybrid} architectures from scratch, interleaving linear and softmax layers. Additionally, models like DeepSeek-V4 \citep{deepseekv4} adopt an intra-layer hybrid of compressed sparse and sliding window attention during pretraining to achieve acceleration. While effective, training these foundation models from scratch requires thousands of GPU hours and massive compute clusters, making such architectures inaccessible to most researchers and practitioners. Furthermore, this inter-layer strategy still retains quadratic layers and fails to leverage the rich, full-attention features already established in existing open-source models.}

\begin{figure*}[t]
    \centering
    \centerline{\includegraphics[width=0.9\textwidth]{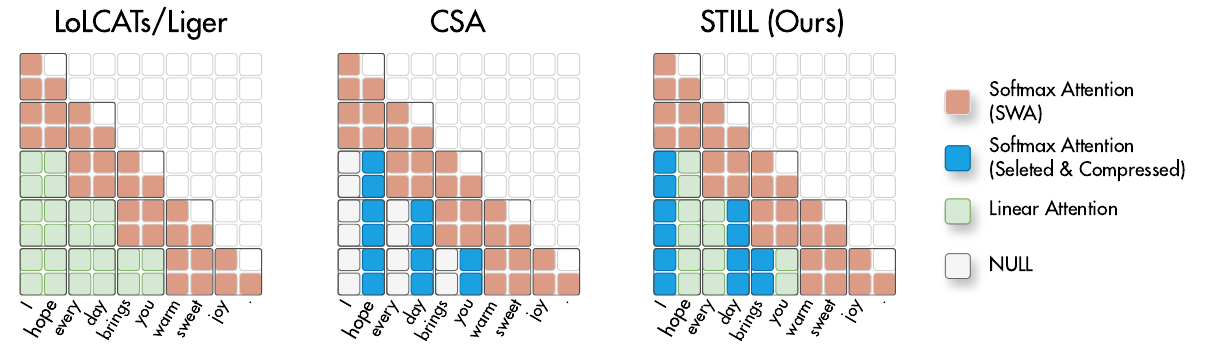}}
    \vspace{-1.5ex}
    \caption{
      \textcolor{black}{\textbf{Illustration of Intra-Layer Hybrid Attention.} While both act as intra-layer hybrids using local Softmax Attention (SA), they differ in handling the remaining sequence. LoLCATs/Liger applies uniform Linear Attention (LA), which limits long-context capacity, while compressed sparse attention (CSA) utilizes sparse compressed SA but completely discards unselected tokens (NULL). STILL selects tokens for high-fidelity SA and preserves the context via LA without information loss.}
    }
    \label{fig:weight}
    \vspace{-3ex}
\end{figure*}

\textcolor{black}{To bypass the prohibitive costs of pretraining, recent work has shifted toward \textit{\textbf{intra-layer} hybrid} attention designs that inherit checkpoints from \textit{already-pretrained} LLMs.}
Approaches such as LoLCATs \citep{lolcats}, Liger \citep{liger} and KL-LS \citep{KLSelecting} distill and fine-tune pretrained LLMs while replacing full softmax attention with intra-layer operator: softmax attention is computed over a fixed-length local window and linear attention summarizes the remaining context, yielding near-linear cost with relatively modest additional training. 
Most of these pipelines follow a similar recipe: start from a pretrained LLM, route the context tokens into SA branch and LA branch to encode the current token, train the learnable LA branch on a restricted subset of tokens, and finally finetune the model to align the hybrid operator with the original model behavior. 
This makes linearization feasible \textit{without} rebuilding the model from scratch.

Despite the progress, this paradigm faces two key challenges. 
\textcolor{black}{First, current token routing relies heavily on absolute position (e.g., a sliding window), implicitly assuming key dependencies are local (Fig. \ref{fig:weight}). 
For the remaining global context, approaches such as LoLCATs \citep{lolcats} and Liger \citep{liger} apply uniform linear attention, while Compressed Sparse Attention (CSA) \citep{deepseekv4} condenses tokens into block entries and restricts its attention to a top-$k$ subset, leaving the unselected blocks unattended.
Since crucial evidence often appears far outside this fixed window, pure position-based routing can inadvertently exclude vital tokens from high-fidelity softmax attention.}
Second, linearizing a pretrained Transformer must respect its distributional geometry, particularly the norm structure induced by softmax attention. Prior work \citep{nalaformer,inline,mala} highlights that pretrained attention is ``norm-aware'': vector magnitudes modulate representational intensity and influence attention weights. However, many linearization schemes introduce learnable feature maps (often implemented with MLPs) that inadvertently \textit{distort} these norms \citep{hedgehog}, so that even a nominally norm-aware kernel may operate on \textit{miscalibrated} magnitudes. Together, these issues suggest that effective linearization requires (i) routing that reflects token importance, not just position, and (ii) kernels that preserve pretrained norm statistics rather than relearning them.


In this work, we propose STILL, a novel linearization framework that converts pretrained Transformers into efficient long-context models via in-layer hybrid attention, while explicitly addressing both challenges.
STILL introduces Self-Saliency Score, which computes token importance within a local window to determine its relevance as a historical context.
Notably, Self-Saliency Score exhibits high consistency between local and global contexts, allowing for reliable token selection using only a sliding-window mechanism.
STILL introduces a Self-Saliency Score that estimates token importance using only local computation, yet is empirically consistent with global relevance. This enables reliable selection of a small set of salient tokens for softmax attention, while the remaining context is summarized with linear attention.
In addition, to mitigate the norm distortion during linearization, we propose the Norm-Preserved Feature Map (NP-Map) that decouples direction from magnitude and reinjects pretrained norms, aligning the linear kernel with the model’s representational intensity.
\textcolor{black}{Finally, to overcome the branching overhead and parallelization bottlenecks of naive token-wise routing, STILL introduces a delayed, chunk-level selection strategy. By making routing decisions collectively per chunk, this hardware-friendly approach maximizes throughput, effectively translating theoretical linear complexity into tangible wall-clock acceleration during both training and inference.}

\textcolor{black}{
We comprehensively evaluate STILL across both standalone intra-layer modules and intra-inter layer hybrid frameworks. Remarkably, STILL achieves linear-time scalability using less than 0.4B training tokens. This minimal cost translates to massive efficiency gains, yielding an 80\% average memory reduction and an 83\% prefilling speed-up while smoothly scaling sequences beyond 64K. STILL consistently outperforms existing baselines on reasoning tasks (up to +10.5\% on MMLU \citep{mmlu}) and powerfully restores extended-context capabilities on RULER \citep{ruler} and BABILong \citep{babilong}, recovering 86.2\% of full-attention performance with a minimal softmax token budget. Crucially, STILL overcomes the severe performance collapse that plagues baselines at 16K contexts. While prior methods largely fail, our model maintains robust long-range memory, highlighted by a flawless score of 100 on the 16K S-NIAH-1 benchmark. These results show that self-saliency-based token selection and norm-preserved linear attention can largely recover the long-context capability lost in existing linearization methods, thereby enabling effective linearization of pretrained LLMs in long-context regimes.
}




\section{Preliminaries}
\noindent\textbf{Linearizing Pretrained LLMs with Intra-Layer Hybrid Attention.}
Linearizing pretrained large language models (LLMs) aims to reduce the quadratic complexity of Softmax attention while preserving the representational capacity of pretrained weights. 
Intra-Layer Hybrid Attention provides a practical framework by integrating Softmax attention and Linear Attention within the same Transformer layer, where only a subset of tokens is processed using \textit{Softmax Attention} (SA) and the remaining tokens are routed to \textit{Linear Attention} (LA). 
Formally, the attention output at time step $t$ is obtained by linearly combining the two branches:
\begin{equation}
\mathbf{y}_t
= \alpha\mathbf{y}_t^{\mathrm{SA}}
+ \beta\mathbf{y}_t^{\mathrm{LA}}.
\end{equation}
This design allows the model to retain high-fidelity attention for selected tokens while achieving linear-time complexity for the majority of the sequence. 
When linearizing a pretrained models, Linear Attention is typically initialized by inheriting existing attention weights, followed by lightweight alignment or fine-tuning to mitigate distribution mismatch.
Existing Intra-Layer Hybrid Attention methods commonly adopt Sliding Window Attention (SWA) as a static, position-based routing strategy, where tokens within a fixed local window are processed using Softmax attention to preserve fine-grained local interactions, while tokens outside the window are handled by Linear Attention to capture broader contextual information.

\noindent\textbf{Linear Attention with Learnable Feature Maps.}
Prior works \citep{lolcats,t2r} introduce \emph{learnable feature maps} that enable linear attention to approximate the exponential similarity of softmax attention given a pretrained weight from standard attention. Hedgehog \citep{hedgehog}, as a representative design, employs a lightweight MLP to transform the input and constructs a non-negative feature map by concatenating the softmax activations of the projected features and their negations. However, recent studies \citep{inline,nalaformer,mala} have shown that the exponential function in softmax attention is highly sensitive to vector norms, where the sharpness of the attention distribution is partly controlled by the input magnitude. Consequently, the \emph{norm-aware} property of softmax kernels is semantically valid only when applied to feature norms consistent with those of the pretrained model. However, we observe that in Hedgehog, the MLP transformation preceding the softmax mapping can freely rescale feature norms during fine-tuning, breaking the norm consistency before and after the MLP and causing the subsequent softmax to operate on a distorted norm distribution.


\section{Methodology}
In the following, we present the STILL model.
We first propose the Self-Saliency Score for In-Layer Hybrid Linear Attention, which determines whether long-range dependencies should be modeled by softmax attention or linear attention, avoiding coarse sliding-window-based partitioning.
Then, we present a Norm-preserved Feature Map that explicitly retains feature norm information, facilitating better alignment of linear attention with the inherited model.
Finally, we present the complete STILL model with a chunk-wise parallel implementation that accelerates both training and inference.

\begin{figure*}[t]
    \centering
    \centerline{\includegraphics[width=1\textwidth]{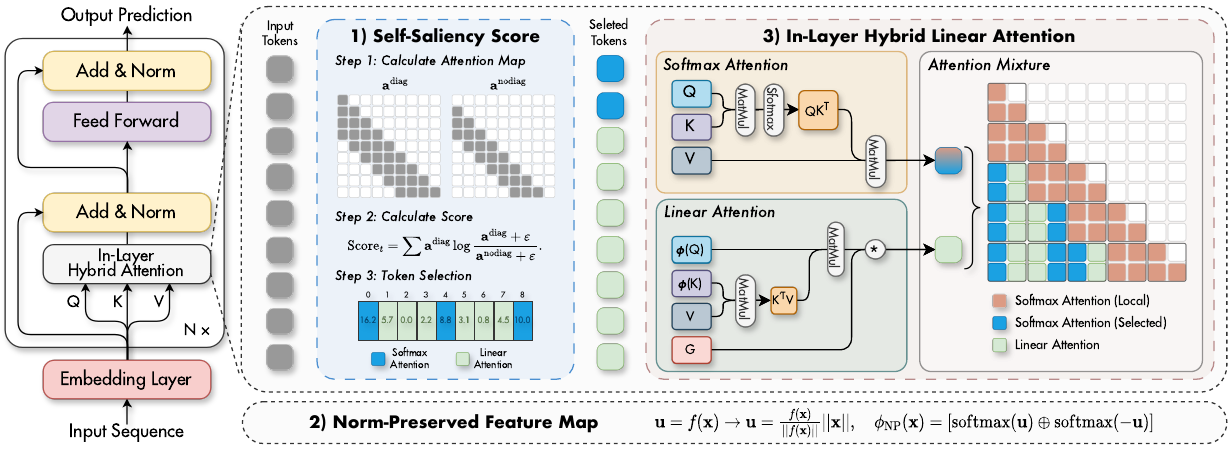}}
    \vspace{-1ex}
    \caption{
      \textbf{Architecture of STILL for Intra-Layer Hybrid Linear Attention.}
      The diagram shows attention computation, chunk-level routing, and hybrid aggregation to outputs. STILL first computes the Self-Saliency Score within sliding-window attention for each token. Top-scoring tokens are routed to the softmax attention, while remaining tokens are processed via linear attention. Chunk-wise selecting replaces per-token selecting, enabling parallel training and inference.
    }
    \vspace{-2ex}
    \label{fig:main_fig}
\end{figure*}

\subsection{Selecting Tokens by Self-Saliency Score}
\label{sec:3-1}
The core of in-layer hybrid attention lies in how to route the past context tokens into softmax attention (SA) and linear attention (LA) when computing the attention of the current token.
Existing in-layer hybrid attention methods typically rely on sliding-window heuristics to route tokens to SA, which are inherently \textit{position-based} and therefore fail to capture token-specific importance.
To enable content-aware token selection, we introduce a \emph{Self-Saliency Score}, which is computed for each token based on its attention weights within a local window. 
Formally, given an input sequence of length $N$, we denote the query and key
representations for each head as
$\{\mathbf{q}_t\}_{t=1}^{N}\in\mathbb{R}^{N\times d}$ and
$\{\mathbf{k}_t\}_{t=1}^{N}\in\mathbb{R}^{N\times d}$, where
$\mathbf{q}_t,\mathbf{k}_t\in\mathbb{R}^{d}$.
We consider a sliding-window attention (SWA) with window radius $w$.
For each query token $\mathbf{q}_t$, we define its local window index set as
\begin{equation}
\mathcal{W}_t=\{j \mid \max(1,t-w+1)\le j \le \min(N,t)\}.
\end{equation}
For each token $t$, we measure the sensitivity of its local attention distribution
to the self-attention term by comparing two SWA distributions:
SWA with the diagonal term (Eq.~\eqref{eq:diag}) is defined as,
\begin{align}
    \mathbf{a}^{\mathrm{diag}}_{t,j}=&\;\;\;\frac{\exp(\mathbf{q}_t\mathbf{k}_j^\top)}{\sum\limits_{m\in\mathcal{W}_t}\exp(\mathbf{q}_t\mathbf{k}_m^\top)},\quad j\in\mathcal{W}_t,
    \label{eq:diag}
\end{align}
and SWA without the diagonal term Eq.~\eqref{eq:nodiag} is defined as,
\begin{align}
\mathbf{a}^{\mathrm{nodiag}}_{t,j}=0~~(t=j),~~~~~
    \mathbf{a}^{\mathrm{nodiag}}_{t,j}=
        \dfrac{\exp(\mathbf{q}_t\mathbf{k}_j^\top)}{\sum\limits_{m\in\mathcal{W}_t\setminus\{t\}} \exp(\mathbf{q}_t\mathbf{k}_m^\top)}~~~ (j\in\mathcal{W}_t\setminus\{t\}).
    \label{eq:nodiag}
\end{align}

Finally, we leverage the two attention distributions to compute the \textbf{Self-Saliency Score} as
\begin{equation}
    \mathrm{Score}_t=\sum_{j\in\mathcal{W}_t}
    \mathbf{a}^{\mathrm{diag}}_{t,j}\log\frac{\mathbf{a}^{\mathrm{diag}}_{t,j}+\epsilon}{\mathbf{a}^{\mathrm{nodiag}}_{t,j}+\epsilon}.
    \label{eq:score}
\end{equation}
The Self-Saliency Score is then used to determine whether the token, when serving as a context token outside the local window, should be routed to the SA or LA branch, allowing the routing decision to adapt to token importance rather than solely rely on position. 
A higher Self-Saliency Score indicates that a token relies strongly on its self-attention term and is likely important under full-context attention.
As shown in Fig. \ref{fig:main_fig}, Top-k tokens are routed to the SA branch, while the other tokens are handled by LA branch, enabling efficient and content-aware hybrid attention.

\begin{figure}[t]
    \centering
    \begin{minipage}[t]{0.48\linewidth}
        \vspace{0.2pt} 
        \centering
        \includegraphics[width=0.94\linewidth]{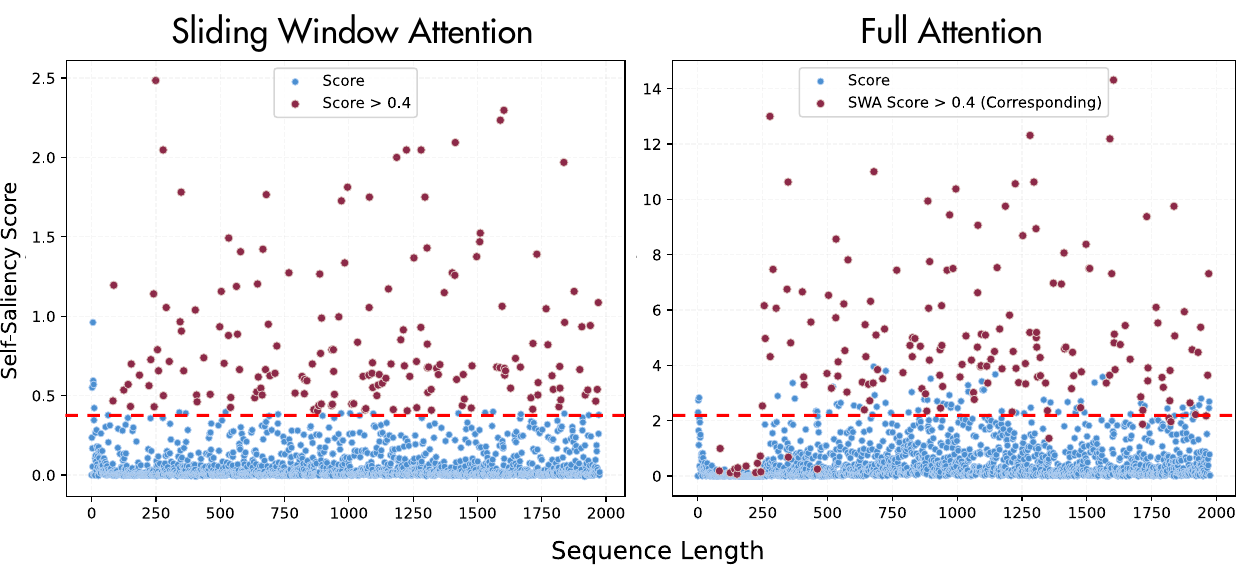}
        \vspace{-2ex}
        \caption{\textbf{Self-Saliency Score Distribution.} Red points mark top-scoring tokens routed to SA under sliding-window attention (left). Highlighting these exact tokens globally (right) shows strong local-global consistency.}
        \label{fig:localvsglobal} 
    \end{minipage}
    \hfill
    \begin{minipage}[t]{0.5\linewidth}
        \vspace{0pt} 
        \centering
        \includegraphics[width=\linewidth]{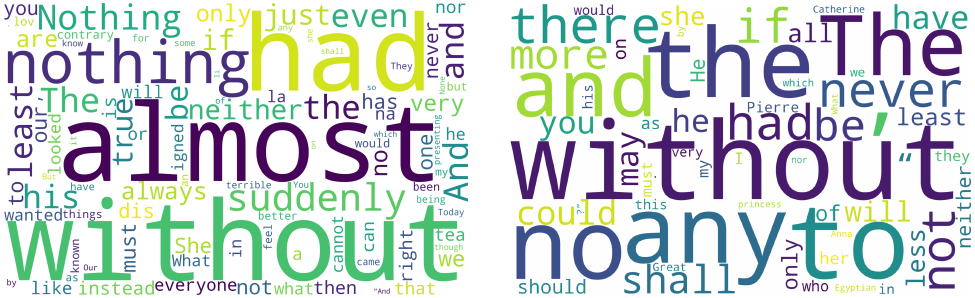}
        \vspace{-1ex}
        \caption{\textbf{Word Cloud of Tokens Most Frequently Selected for the SA Branch by Self-Saliency Score.} The selected set is dominated by \textit{functional} operators (e.g., negation and quantification), suggesting that SA prioritizes precise, context-dependent interactions.}
        \label{fig:wordcloud}
    \end{minipage}
    \vspace{-2ex}
\end{figure}

To evaluate our Self-Saliency Score, we compare score distribution under sliding-window (local) and full-context (global) attention. 
As shown in Fig.~\ref{fig:localvsglobal}, the score distribution is clear: most tokens have near-zero scores, suitable for LA, while only a few reach high scores for SA processing. Importantly, tokens identified as salient locally largely overlap with those selected under full-context attention, demonstrating that global importance can be reliably approximated using only local information. 
\textcolor{black}{\textit{More visualizations across different layers, heads, and window sizes are provided in Appendix~\ref{appendix:morevisualization}, all confirming this conclusion.}}
This consistency enables an efficient, content-aware routing strategy: token saliency is first estimated with sliding-window attention, after which only a small set of globally important tokens is routed to softmax attention, while the majority are handled efficiently by linear attention.
\textcolor{black}{To further justify our token routing strategy, we establish the following theoretical property:}

\begin{proposition}[Local-Global Consistency]
Let $p_{local}$ and $p_{global}$ be the local and global self-attention probabilities. For $\epsilon \to 0^+$, the self-saliency score, $\operatorname{Score}_t$, satisfies:
\begin{enumerate}
    \item \textbf{Monotonicity}: $\frac{\partial \operatorname{Score}_t}{\partial p_{local}} > 0$ for all $p_{local} \in (0, 1)$.
    \item \textbf{Rank Preservation}: $\operatorname{Score}_t$ is rank-preserving with respect to the expected global self-attention probabilities, i.e.:
    \begin{equation}
        \operatorname{Score}_i > \operatorname{Score}_j \Rightarrow \mathbb{E}[p_{global}^{(i)}] > \mathbb{E}[p_{global}^{(j)}].
    \end{equation}
\end{enumerate}
\end{proposition}
\textcolor{black}{Thus, local Top-$\lambda$ selection based on $\operatorname{Score}_t$ is a probabilistically optimal proxy for global routing. Full proofs are in Appendix \ref{appendix:proof}.}

For a qualitative view of what the score prioritizes, we visualize the most frequently selected SA tokens as a word cloud in Fig.~\ref{fig:wordcloud}; the prominence of \textbf{negation} (\textit{not, no, never}), \textbf{modality} (\textit{will, would, may}), and \textbf{quantification} (\textit{all, any, nothing}) tokens supports our hypothesis that these high-scope operators particularly benefit from high-fidelity softmax attention.

\subsection{Norm-Preserved Feature Map (NP-Map)}
\label{sec:3-2}
A critical yet often overlooked challenge in model linearization is the preservation of the norm-aware property intrinsic to the Softmax kernel.
As established in NaLaFormer \citep{nalaformer}, the exponential function in standard attention naturally incorporates vector magnitudes into the attention score, allowing the model to distinguish tokens based on their representational intensity.

While Hedgehog \citep{hedgehog} feature maps employ a softmax-based feature map to approximate this property, they typically introduce an additional MLP before the mapping to transform features.
We observe that this MLP, while flexible, inevitably distorts the pre-trained norm distribution. 
Consequently, even though the subsequent softmax mapping is technically norm-aware, it operates on misaligned norms.

To address this limitation, we propose the \textbf{Norm-Preserved Feature Map} (NP-Map), a novel learnable feature map designed to enforce norm-invariance during the MLP transformation. 
Instead of allowing the MLP to freely scale the features, our approach explicitly decouples the feature's direction from its magnitude, recovering the original pre-trained norm while utilizing the MLP to only refine the directional components.
Formally, for a given input $\mathbf{x}\in \{\mathbf{q,k}\}$ derived from the pre-trained checkpoint, the NP-Map is defined as,
\begin{equation}
    \begin{aligned}
        \mathbf{u}=f(\mathbf{x})\to \mathbf{u}=\frac{f(\mathbf{x})}{\|f(\mathbf{x})\|}\|\mathbf{x}\|,~~~
        \phi_{\mathrm{NP}}(\mathbf{x})={\left[\mathrm{softmax}\left(\mathbf{u}\right)\oplus\mathrm{softmax}\left(-\mathbf{u}\right)\right]},
    \end{aligned}
    \label{eq:feature_map}
\end{equation}
where $f(\cdot)$ represents the learnable MLP as used in Hedgehog.
By re-injecting the original norm $\|\mathbf{x}\|$ into the transformed feature $\mathbf{u}$, NP-Map ensures that the attention mechanism remains strictly aligned with the pre-trained representational intensity. 
This allows the linear model to satisfy the non-negativity constraint while effectively inheriting the fine-grained semantic information from the pretrained Transformer.
Furthermore, as gated attention \citep{gla,gateattn} effectively enhances the rank of linear attention while maintaining linear complexity with minimal computational overhead, we incorporate an additional gating mechanism. The complete linear attention computation is formulated as follows:
\begin{align}
    \begin{aligned}
        \mathbf{y}_t &=\frac{\phi_{\text{NP}}(\mathbf{q}_t)\sum_{i=1}^{t}\phi_{\text{NP}}(\mathbf{k}_i)^\top\mathbf{v}_i}{\phi_{\text{NP}}(\mathbf{q}_t)\sum_{j=1}^{t}\phi_{\text{NP}}(\mathbf{k}_j)^\top}\odot\mathbf{g}_t.
    \end{aligned}
    \label{eq:gla}
\end{align}


\subsection{STILL: Selecting Tokens for In-Layer Hybrid Linear Attention to Linearize LLMs}
\label{sec:3-3}
In this section, we propose \textbf{STILL}, an intra-layer hybrid linear attention method that combines self-saliency-based token selection with a norm-preserved learnable feature map. 
As illustrated in Fig.~\ref{fig:main_fig}(a), STILL maintains a fixed-budget
Softmax branch for recent and salient historical tokens, while the remaining
historical context is summarized by linear attention. The same routing rule is
used for recurrent decoding and chunk-wise parallel computation during training
and prefilling.

\noindent\textbf{Fixed-Budget Routing.}
To ensure bounded memory and linear complexity, STILL adopts a
configuration-level fixed Softmax KV-cache budget $m$ per attention head, which is
independent of the runtime sequence length $N$. Among the $m$ cache slots, $W$ are reserved
for the most recent tokens in the sliding window, while the remaining $m-W$
slots are allocated to salient historical tokens.
Given a configured maximum context length $N_{\max}$ and chunk size $C$, we set
\begin{equation}
    T_{\max}=\left\lceil N_{\max}/C \right\rceil,
    \qquad
    \lambda=
    \left\lfloor (m-W)/{T_{\max}}\right\rfloor,
    \label{eq:fixed_budget}
\end{equation}
where $\lambda$ denotes the number of salient tokens retained from each
completed chunk. For any sequence length $N\leq N_{\max}$ with
$T=\lceil N/C\rceil$ completed chunks, the number of historical salient tokens
is bounded by $T\lambda \leq T_{\max}\lambda \leq m-W$.

Therefore, the total Softmax cache, including both the $W$ local tokens and
the historical salient tokens, never exceeds $m$ tokens per head.
All routing decisions are made independently per attention head, with each head selecting and retaining its own salient tokens; we omit head indices
for brevity.
Once selected, a salient token is retained permanently in the SA cache
and is neither rescored nor evicted; unselected tokens are instead accumulated
into the recurrent LA state.

\noindent\textbf{Recurrent Form.}
In the decoding stage, STILL retain a small sliding-window attention for recent tokens to preserve local context.
When a token moves out of the sliding window, we aim to the self-saliency score to decide whether it is routed to SA or LA.
However, directly performing this routing decision at each decoding step would incur non-negligible computational overhead and low parallelism.
To improve decoding efficiency, we therefore propose a delayed selection strategy, where token routing is performed at the granularity of chunks rather than at each time step.

We index the decoding step as $t = (a+1)T + b$, where $T$ denotes the chunk size, $a$ is the index of the second latest completed chunk, and $b$ indicates the position of the current token within the ongoing chunk.
At each decoding step, we compute the self-saliency score for the current token, as defined in Eq.~\eqref{eq:score}.
When $b = 1$, we rank the tokens in the most recently completed chunk $a$ according to their S3 scores and select the top-$\lambda$ tokens as \textit{SA tokens}.
\begin{equation}
    \begin{aligned}
        \mathbf{M}_{a}^{\mathcal{SA}}
=\operatorname{Top}_{\lambda}\!\left(\mathrm{Score}_{a[1]:a[T]}\right),~~~
\mathbf{M}_{a}^{\mathcal{LA}}=\{1,\dots,T\} \setminus \mathbf{M}_{a}^{\mathcal{SA}},
    \end{aligned}
    \label{eq:delayed_selection}
\end{equation}
where $\mathbf{M}_{a}^{\mathcal{SA}}$ denotes the indices of the top-$\lambda$ scores, and 
$\mathbf{M}_{a}^{\mathcal{LA}}$ denotes the complementary set of indices.
Based on the selection results, tokens are divided into \textit{salient} and \textit{unsalient} ones.
High-scoring and locally relevant key–value pairs are appended to the salient cache $\mathbf{h}_t$.
The remaining tokens are written into the KV state $({\mathbf{s}}_t, {\mathbf{z}}_t)$ and processed by the linear attention branch.
Formally, we have:
\begin{equation}
    \begin{aligned}
        \mathbf{h}_{t}
        =&\{(\mathbf{k}_{i},\mathbf{v}_{i}) \mid
        \max(0,aT)< i \le \min(N,(a+1)T+b)\}\\
        &\cup
        \begin{cases}
            \mathbf{h}_{t-1}\cup
            \Bigl\{(\mathbf{k}_{a[i]},\mathbf{v}_{a[i]})
            \mid i\in\mathbf{M}_{a}^{\mathcal{SA}}\Bigr\},
            & a\geq1,b=1,\\
            \varnothing, & a<1,\\
            \mathbf{h}_{t-1}, & o.w.,
        \end{cases}
    \end{aligned}
    \label{eq:cache_routing_sum}
\end{equation}

\vspace{-2ex}
\begin{equation}
    \mathbf{s}_{t} =
    \begin{cases}
        \mathbf{s}_{t-1}
        +\displaystyle\sum_{i\in\mathbf{M}_{a}^{\mathcal{LA}}}
        (\phi_{\mathrm{NP}}(\mathbf{k}_{a[i]}))^\top
        \mathbf{v}_{a[i]},
        & a\geq1,b=1,\\
        \mathbf{0}, & a<1,\\
        \mathbf{s}_{t-1}, & o.w.,
    \end{cases}
\end{equation}
\begin{equation}
    \mathbf{z}_{t} =
    \begin{cases}
        \mathbf{z}_{t-1}
        +\displaystyle\sum_{i\in\mathbf{M}_{a}^{\mathcal{LA}}}
        (\phi_{\mathrm{NP}}(\mathbf{k}_{a[i]}))^\top,
        & a\geq1,b=1,\\
        \mathbf{0}, & a<1,\\
        \mathbf{z}_{t-1}, & o.w..
    \end{cases}
\end{equation}
Finally, computes the in-layer hybrid linear attention output token $\mathbf{y}_t$ as
\begin{equation}
    \mathbf{y}_t = \frac{\mathcal{N}_\mathrm{SA} +\mathcal{N}_\mathrm{LA}}{\mathcal{D}_\mathrm{SA}+\mathcal{D}_\mathrm{LA}},
\end{equation}
\begin{equation}
    \begin{aligned}
        \mathrm{SA:}&
        \begin{cases}
            \mathcal{N}_\mathrm{SA} = \sum_{(\mathbf{\tilde{k}},\mathbf{\tilde{v}})\in \mathbf{h}_t}\exp(\mathbf{q}_t\mathbf{\tilde{k}}^\top)\tilde{\mathbf{v}}, \\
            \mathcal{D}_\mathrm{SA} = \sum_{(\mathbf{\tilde{k}},\mathbf{\tilde{v}})\in \mathbf{h}_t}\exp(\mathbf{q}_t\mathbf{\tilde{k}}^\top),
        \end{cases}
        \;
        \mathrm{LA:}&
        \begin{cases}
            \mathcal{N}_\mathrm{LA} = \phi_\mathrm{NP}(\mathbf{q}_t)\mathbf{s}_t\odot \mathbf{g}_t, \\
            \mathcal{D}_\mathrm{LA} = \phi_\mathrm{NP}(\mathbf{q}_t)\mathbf{z}_t.
        \end{cases}
    \end{aligned}
\end{equation}

In this way, STILL effectively routes appropriate tokens to softmax attention (SA) and linear attention (LA), leveraging the complementary strengths of both mechanisms within a unified computation.

\noindent\textbf{Chunk-Wise Parallel Form.}
To improve parallelism during training and pre-filling while maintaining a unified training--inference formulation, our method can be implemented in a chunk-wise parallel manner.
We divide the input sequence of length $N$ into $T=\lceil N/C \rceil$ chunks, each containing $C$ tokens.
For each chunk, we first compute the self-saliency scores for all tokens and perform a local--global consistent selection within the chunk based on these scores.
The selected tokens are then routed to softmax attention (SA), while the remaining tokens are assigned to linear attention (LA).
Both SA and LA are computed efficiently in a chunk-wise parallel fashion, and their outputs are finally combined to produce the layer output.
This chunk-wise formulation enables parallel computation within each chunk while preserving the same routing and memory-update behavior as in decoding.
The overall procedure is summarized in Appendix \ref{appendix:algorithm}.

\noindent\textbf{Complexity Analysis.}
The softmax attention branch requires $\mathcal{O}(Nmd)$ computations where $m$ represents the salient tokens size.
The linear attention branch processes the remaining tokens with a complexity of $\mathcal{O}((N - m)dd')$.
Furthermore, the delayed selection strategy contributes $\mathcal{O}(T\log(C))$ to the total complexity. 
Summarizing these components, the overall complexity of STILL is given in Eq.~\eqref{eq:complexity}, confirming that it scales linearly with the sequence length $N$.

\vspace{-2ex}
\begin{equation}
    \begin{aligned}
        \underbrace{\mathcal{O}(Nmd)}_{\text{SA}}
        +\underbrace{\mathcal{O}((N - m)dd')}_{\text{LA}}
        +\underbrace{\mathcal{O}(T\log(C))}_{\text{Selecting}}
        =\mathcal{O}(N)\quad (\text{Total}).
    \end{aligned}
    \label{eq:complexity}
\end{equation}
\vspace{-2ex}

\section{Experiments}

We evaluate STILL across three critical dimensions. We first assess its representational capacity on standard reasoning and challenging long-context tasks, including a demonstration of its compatibility within an intra-inter hybrid framework. Next, we quantify its hardware efficiency via memory and latency analyses. Finally, we validate its structural robustness across diverse backbone architectures, with detailed ablation studies deferred to Appendix~\ref{appendix:ablation}.

\subsection{Experimental Setups}
\noindent\textbf{Training Strategy and Datasets.} Following previous work \citep{lolcats}, our training strategy employs a two-stage pipeline consisting of Attention Transfer and Low-rank Linearization, which including two stages. Stage 1 freezes the teacher and trains the learnable feature/gating projections in NP-Map via MSE. Stage 2 performs end-to-end LoRA \citep{lora} fine-tuning. Both stages are conducted on the Cleaned Alpaca dataset \citep{alpaca} for 20M tokens with a sequence length of 1024.

\noindent\textbf{Evaluation Settings.}
We evaluate all models using the lm-evaluation-harness framework \citep{lmeval}, following its standard evaluation protocols.
All models are evaluated without task-specific finetuning.
For each benchmark, we report the primary metric defined by the evaluation harness, such as accuracy or normalized accuracy, and ensure consistent evaluation settings across all compared methods.

\subsection{Commonsense and General Reasoning Tasks}
\noindent\textbf{Settings.}
\noindent\textbf{Settings.} We evaluate STILL's commonsense and general reasoning on PIQA~\citep{piqa}, ARC-E/C~\citep{arc}, HellaSwag~\citep{hellaswag}, WinoGrande~\citep{winogrande}, and MMLU~\citep{mmlu} under standard zero-shot settings (five-shot for MMLU)~\citep{supra,lolcats,liger}. Using Llama 3 and 3.1 8B~\citep{llamba} as backbones, we compare against original Transformers, from-scratch subquadratic, inter-layer hybrid, and linearized baselines.

\noindent\textbf{Results.}
Tab.~\ref{tab:commonsense-reasoning} shows STILL achieves the best average performance (with and without MMLU), consistently outperforming linearized baselines. Notably, under the Llama 3.1 8B backbone, STILL improves MMLU accuracy over Liger~\citep{liger} from 46.9 to 61.3, strongly preserving the backbone’s reasoning capabilities, and impressively surpasses the original full-attention model's average performance by 1.5.
Extended baseline comparisons and results across various pretrained LLM backbones are provided in Appendix Tab.~\ref{tab:teacher} and Tab.~\ref{tab:commonsense-reasoning_full}, respectively.

\begin{table}[t]
    \caption{\textbf{Comparison of the Commonsense and General Reasoning Tasks}.
    Comparison of STILL with baseline methods under Llama 3 8B and Llama 3.1 8B models. STILL consistently demonstrates strong performance across benchmarks.
    }
    \vspace{-1ex}
    \label{tab:commonsense-reasoning}
    \begin{center}
            \resizebox{1\linewidth}{!}{
                \begin{tabular}{l|c|ccccc|c|cc}
                    \toprule
                    \multirow{2}{*}{\textbf{\textsc{Model}}}
                    & \textbf{\textsc{Training}}
                    & \textbf{\textsc{PIQA}}
                    & \textbf{\textsc{ARC-e}}
                    & \textbf{\textsc{ARC-c}}
                    & \textbf{\textsc{Hella.}}
                    & \textbf{\textsc{Wino.}}
                    & \textbf{\textsc{MMLU}}
                    & \textbf{\textsc{Avg.}}
                    & \textbf{\textsc{Avg.}}
                    \\
                    & \textbf{\textsc{Tokens}}
                    & acc $\uparrow$
                    & acc $\uparrow$ 
                    & acc$_n$ $\uparrow$
                    & acc$_n$ $\uparrow$  
                    & acc $\uparrow$  
                    & (5-shot)  $\uparrow$  
                    & (w. MMLU)
                    & (wo. MMLU)
                    \\
                    \midrule 
                    \rowcolor[HTML]{F1E9E7}
                    \multicolumn{10}{l}{\textbf{\textit{Transformer}}}
                    \\
                    Llama 3 8B~\citep{llama}
                    & 15000$+$
                    & 79.5
                    & 80.0
                    & 53.3
                    & 79.1
                    & 73.1
                    & 65.3
                    & 71.8
                    & 73.0
                    \\
                    Llama 3.1 8B~\citep{llama}
                    & 15000$+$
                    & 81.1
                    & 81.7
                    & 55.1
                    & 79.3
                    & 73.9
                    & 68.0
                    & 74.2
                    & 73.2
                    \\
                    \midrule 
                    \rowcolor[HTML]{F1E9E7}
                    \multicolumn{10}{l}{\textbf{\textit{Subquadratic}}}
                    \\
                    Mamba2 8B~\citep{mamba2}
                    & 3500
                    & 79.8
                    & 75.9
                    & 48.1
                    & 77.1
                    & 71.6
                    & 48.7
                    & 67.0
                    & 70.6
                    \\
                    Griffin 7B~\citep{griffin}
                    & 300
                    & 81.0
                    & 75.4
                    & 47.9
                    & 78.6
                    & 72.6
                    & 39.3
                    & 65.8
                    & 71.1
                    \\
                    Falcon3-Mamba 7B~\citep{falcon}
                    & 7300
                    & 80.6
                    & 78.9
                    & 57.1
                    & 80.1
                    & 73.7
                    & 55.1
                    & 70.9
                    & 74.1
                    \\
                    \midrule
                    \rowcolor[HTML]{F1E9E7}
                    \multicolumn{10}{l}{\textbf{\textit{Hybird}}}
                    \\
                    StripedHyena-Nous 7B~\citep{stripedhyena}
                    & -
                    & 78.8
                    & 77.2
                    & 40.0
                    & 76.4
                    & 66.4
                    & 26.0
                    & 60.8
                    & 67.8
                    \\
                    Zamba 7B~\citep{zamba}
                    & 1000
                    & 81.4
                    & 74.5
                    & 46.6
                    & 80.2
                    & 76.4
                    & 57.7
                    & 69.5
                    & 71.8
                    \\
                    RecurrentGemma 9B~\citep{recurrentgemma}
                    & 2000
                    & 80.6
                    & 78.9
                    & 57.1
                    & 80.1
                    & 73.7
                    & 55.1
                    & 70.9
                    & 74.1
                    \\
                    \midrule
                    \rowcolor[HTML]{F1E9E7}
                    \multicolumn{10}{l}{\textbf{\textit{Linearized from Llama 3 8B}}}
                    \\
                    T2R-LoLCATs~\citep{t2r}
                    & 0.04
                    & 62.0
                    & 42.1
                    & 24.7
                    & 32.7
                    & 48.3
                    & 23.2
                    & 38.8
                    & 42.0
                    \\
                    Hedgehog-LoLCATs~\citep{hedgehog}
                    & 0.04
                    & 77.4
                    & 71.1
                    & 40.6
                    & 50.7
                    & 54.3
                    & 24.2
                    & 53.1
                    & 58.8
                    \\
                    SUPRA~\citep{supra}
                    & 20
                    & 78.9
                    & 75.1
                    & 46.5
                    & 71.7
                    & 65.8
                    & 40.9
                    & 63.2
                    & 67.6
                    \\
                    LoLCATs~\citep{lolcats}
                    & 0.04
                    & 80.0
                    & 78.8
                    & 52.2
                    & 75.6
                    & \textbf{73.3}
                    & \underline{43.9}
                    & 67.3
                    & 72.0
                    \\ 
                    Liger-GLA~\citep{liger}
                    & 0.02
                    & \underline{80.3}
                    & \underline{81.1}
                    & \underline{52.5}
                    & \underline{76.3}
                    & 72.0
                    & 43.4
                    & \underline{67.6}
                    & \underline{72.4}
                    \\
                    \rowcolor[HTML]{E2ECF9}
                    STILL~(Ours)
                    & 0.04
                    & \textbf{81.3}
                    & \textbf{81.7}
                    & \textbf{54.8}
                    & \textbf{78.1}
                    & \underline{73.2}
                    & \textbf{59.7}
                    & \textbf{71.5}
                    & \textbf{73.8}
                    \\ 
                    \midrule
                    \rowcolor[HTML]{F1E9E7} 
                    \multicolumn{10}{l}{\textbf{\textit{Linearized from Llama 3.1 8B}}}
                    \\
                    Llamba~\citep{llamba}
                    & 12
                    & \underline{80.9}
                    & 82.5
                    & 54.6
                    & \underline{77.6}
                    & 73.3
                    & \underline{60.0}
                    & \underline{71.5}
                    & \underline{73.8}
                    \\
                    LoLCATs~\citep{lolcats}
                    & 0.04
                    & \underline{80.9}
                    & \underline{82.6}
                    & 54.7
                    & 76.6
                    & \textbf{73.6}
                    & 50.8
                    & 69.9
                    & 73.7
                    \\ 
                    Liger-GLA~\citep{liger}
                    & 0.02
                    & 80.5
                    & 81.8
                    & \underline{55.6}
                    & 75.4
                    & 69.5
                    & 46.9
                    & 68.3
                    & 72.6
                    \\
                    \rowcolor[HTML]{E2ECF9}
                    STILL~(Ours)
                    & 0.04
                    & \textbf{81.3}
                    & \textbf{83.0}
                    & \textbf{56.7}
                    & \textbf{79.0}
                    & \underline{73.4}
                    & \textbf{61.3}
                    & \textbf{72.5}
                    & \textbf{74.7}
                    \\ 
                    \bottomrule
            \end{tabular}
        }
    \end{center}
\end{table}


\subsection{Long-Context Benchmarks}
To evaluate the capability of STILL in handling extended dependencies, we conduct experiments on two challenging long-context tasks, RULER \citep{ruler} and BABILong \citep{babilong}. These benchmarks assess the model's ability to maintain performance as the sequence length increases.

\subsubsection{RULER Benchmark}
We firstly evaluate STILL's long-context retrieval and memory capabilities on the RULER benchmark, covering both \textit{Single Needle-in-a-Haystack (S-NIAH)} retrieval tasks and the \textit{extended RULER} benchmark that requires stronger memory and state tracking.

\noindent\textbf{S-NIAH Settings.} We evaluate S-NIAH \citep{ruler} up to 4K context lengths using the Llama 3.1 8B as the pretrained backbone. We compare STILL against sliding window attention (SWA) only, LoLCATs \citep{lolcats} and Liger \citep{liger} (window sizes: 128, 256, 512). To match baseline cached tokens, STILL keeps the total cache (selected tokens + sliding-window) at the baseline window size.

\noindent\textbf{S-NIAH Results.}
Tab.~\ref{tab:niah_single} shows STILL consistently outperforms SWA and linearized baselines under equal or smaller cache budgets, with the gap widening at longer contexts where these baselines largely fail.
Notably, at 4K context length, STILL achieves 86.2\% on S-NIAH, whereas LoLCATs and Liger largely fail.
Extended model configurations and comprehensive results for the NIAH task are detailed in Appendix Tab.~\ref{tab:niah_single_full}.

\noindent\textbf{Full RULER Settings.}
We evaluate STILL on the extended RULER benchmark~\citep{ruler} at a 4K context length across eight sub-tasks requiring stronger memory and state tracking. Using the Llama 3.1 8B backbone, we compare against Mamba2~\citep{mamba2}, LoLCATs~\citep{lolcats}, and Liger~\citep{liger}, with the latter two using an 896 window size. STILL operates with a 128 window size and a strict 896 total token cap.

\noindent\textbf{Full RULER Results.}
Tab.~\ref{tab:ruler} shows that STILL achieves the highest overall average score of 47.9, exceeding LoLCATs and Liger by a large margin.
These tasks typically require understanding the entire context, indicating that STILL exhibits stronger long-context memory and state-tracking capabilities under tight computational constraints.

\begin{table}[ht]
    \centering
    \begin{minipage}[t]{0.44\linewidth}
    \centering
    \begin{small}
    \caption{\textbf{Comparison of the Single Needle-in-a-Haystack (S-NIAH) Tasks}. Accuracy across 0.5K–4K contexts under the same cache budgets.}
    \vspace{-1ex}
    \setlength{\tabcolsep}{8pt}
    \label{tab:niah_single}
    \resizebox{1\linewidth}{!}
    {
        \begin{tabular}{l|cccc}
            \toprule
            \textbf{\textsc{S-NIAH}}
            & 0.5K 
            & 1K 
            & 2K 
            & 4K 
            \\
            \midrule
            SWA & 100 & 25.2 & 14.6 & 5.8 \\
            LoLCATs~\citep{lolcats}& 100 & 65.6 & 24.6 & 8.8 \\
            Liger-GLA~\citep{liger}& 73.4 & 1.0 & 0.0 & 0.0 \\
            \rowcolor[HTML]{E2ECF9}  
            STILL~(Ours) & \textbf{100} & \textbf{99.6} & \textbf{89.0} & \textbf{86.2} \\
            \rowcolor[HTML]{E2ECF9}  
            ~~~~~---\textit{Improved} & \textcolor{blue}{+0.0} & \textcolor{blue}{+34.0} & \textcolor{blue}{+64.4} & \textcolor{blue}{+77.4}  \\
            \bottomrule
        \end{tabular}
    }
    \end{small}
    \end{minipage}
    \hfill
    \begin{minipage}[t]{0.55\linewidth}
    \centering
    \begin{small}
    \caption{\textbf{Full RULER Benchmark Results.} Comprehensive performance comparison at a 4K context length, evaluating long-range memory and state tracking abilities across multiple tasks.}
    \vspace{-1ex}
    \label{tab:ruler}
    \setlength{\tabcolsep}{1.2pt}
    \resizebox{1\linewidth}{!}{
        \begin{tabular}{l|cccccccc|c}
            \toprule
            \textbf{\textsc{Model}} & \textbf{\textsc{MK}}& \textbf{\textsc{MQ}}& \textbf{\textsc{MV}} & \textbf{\textsc{CWE}} & \textbf{\textsc{FWE}} & \textbf{\textsc{HQA}} & \textbf{\textsc{SQA}} & \textbf{\textsc{VT}} & \textbf{\textsc{Avg.}} \\
            \midrule
            Mamba2 \citep{mamba2} & 19.9 & 49.1 & 35.0 & 32.9 & \textbf{76.6} & 31.8 & 35.5 & \textbf{76.5} & 44.7\\
            LoLCATs \citep{lolcats} & 4.9 & 3.3 & 3.6 & 3.0 & 13.6 & 14.2 & 14.0 & 0.7 & 7.2\\
            Liger-GLA \citep{liger} & 0 & 0 & 0 & 0.2 & 0.7 & 6.4 & 9.1 & 0 & 2.1 \\
            \rowcolor[HTML]{E2ECF9}
            STILL (Ours) & \textbf{21.5} & \textbf{52.2} & \textbf{51.3} & \textbf{36.7} & 76.3 & \textbf{32.2} & \textbf{49.6} & 63.7 & \textbf{47.9}\\
            \bottomrule
        \end{tabular}
    }
    \end{small}
    \end{minipage}
    \vspace{-3ex}
\end{table}

\subsubsection{BABILong Benchmark}


\noindent\textbf{Settings.} We evaluate long-context reasoning on BABILong \citep{babilong}, which extends bAbI \citep{babi} QA by embedding facts within long irrelevant contexts to test distant evidence integration. Using the Llama 3.1 8B backbone, we compare STILL against SWA and LoLCATs across tasks QA1--QA5 under a strict 256-token total cache budget.

\noindent\textbf{Results.}
Tab.~\ref{tab:babilong} shows that STILL consistently achieves the best accuracy across all context lengths. At 4K, STILL improves QA1 from 3 for LoLCATs and 1 for SWA to 10, and improves QA5 from 6 for both LoLCATs and SWA to 24, demonstrating stronger long-context reasoning and evidence aggregation. Full results are provided in Appendix Tab. \ref{tab:babilong_full}.

\begin{table}[ht]
    \centering
    \begin{small}
    \setlength{\tabcolsep}{4pt}
    \caption{\textbf{BABILong Benchmark Results.}
    Performance comparison on BABILong across increasing context lengths (0K$\sim$4K), evaluating long-context reasoning performance.
    }
    \vspace{-1ex}
    \label{tab:babilong}
    \resizebox{1\linewidth}{!}{
        \begin{tabular}{l|llll|llll|llll}
            \toprule
            \multirow{2}{*}{\textbf{\textsc{Model}}}
            & \multicolumn{4}{c|}{\textbf{\textsc{QA1}}}
            & \multicolumn{4}{c|}{\textbf{\textsc{QA4}}}
            & \multicolumn{4}{c}{\textbf{\textsc{QA5}}}
            \\
            & \;0K 
            & 1K 
            & 2K 
            & 4K 
            & \;0K 
            & 1K 
            & 2K 
            & 4K 
            & \;0K 
            & 1K 
            & 2K 
            & 4K 
            \\
            \midrule 
            SWA & \;93 & 2 & 1 & 1 &\;74&0&0&0& \;68 & 18 & 11 & 6 \\
            LoLCATs \citep{lolcats} & \;100 & 22 & 5 & 3 &\;65&21&9&1& \;62 & 42 & 17 & 6 \\
            \rowcolor[HTML]{E2ECF9}
            STILL (Ours) & \;\textbf{100}$_{\textcolor{blue}{+0}}$ & \textbf{45}$_{\textcolor{blue}{+23}}$ & \textbf{22}$_{\textcolor{blue}{+17}}$ & \textbf{10}$_{\textcolor{blue}{+7}}$ &\;\textbf{82}$_{\textcolor{blue}{+8}}$&\textbf{59}$_{\textcolor{blue}{+28}}$&\textbf{30}$_{\textcolor{blue}{+21}}$&\textbf{9}$_{\textcolor{blue}{+8}}$& \;\textbf{83}$_{\textcolor{blue}{+15}}$ & \textbf{77}$_{\textcolor{blue}{+35}}$ & \textbf{45}$_{\textcolor{blue}{+28}}$ & \textbf{24}$_{\textcolor{blue}{+18}}$ \\
            \bottomrule
        \end{tabular}
    }
    \end{small}
\end{table}

\subsection{Architectural Compatibility Analysis: Intra-Inter Layer Hybrid Strategy}
\label{sec:exp_hierarchical_hybrid}
\noindent\textbf{Settings.}
To further evaluate architectural compatibility, we integrate our intra-layer mechanism with the KL-LS inter-layer strategy~\citep{KLSelecting} on Llama 3.2 1B~\citep{llama}. Specifically, following KL-LS, 25\% of the layers are retained as full softmax attention, while in the remaining layers, STILL selects 12.5\% of the tokens for local softmax attention. This proportional token budget is used only in this compatibility experiment to maintain a meaningful long-context comparison, rather than in the fixed-budget setting underlying our linear-complexity claim. We train the model on the DCLM~\citep{dclm}: 100M tokens at 512 context, followed by 250M tokens at 4K context.

\textcolor{black}{
\noindent\textbf{Results.} 
As shown in Tab. \ref{tab:ruler-kl}, STILL demonstrates absolute superiority on the RULER benchmark. STILL perfectly solves S-NIAH-1 up to 16K with a score of 100, and shows remarkable robustness on S-NIAH-2 at 16K, surpassing LoLCATs by 65.4. Consequently, STILL yields a RULER score of 27.20 at 16K, outperforming Zebra-Llama \citep{zebrallama} by 13.83, demonstrating its strong architectural compatibility within intra-layer hybrid frameworks.
}


        

\begin{table}[ht]
    \centering
    \begin{small}
    \caption{\textbf{Intra-Inter Layer Hybrid Results on the RULER Benchmark.} All models are linearized from Llama 3.2 and evaluated over context lengths ranging from 4K to 16K.}
    \vspace{-1ex}
    \label{tab:ruler-kl}
    \resizebox{1\linewidth}{!}
    {
        \begin{tabular}{l|ccc|ccc|ccc|ccc}
            \toprule
            \multirow{2}{*}{\textbf{\textsc{Model}}}
            & \multicolumn{3}{c|}{\textbf{\textsc{S-NIAH-1}}}
            & \multicolumn{3}{c|}{\textbf{\textsc{S-NIAH-2}}}
            & \multicolumn{3}{c|}{\textbf{\textsc{S-NIAH-3}}}
            & \multicolumn{3}{c}{\textbf{\textsc{RULER}}}
            \\
            & 4K 
            & 8K 
            & 16K 
            & 4K 
            & 8K 
            & 16K 
            & 4K 
            & 8K 
            & 16K 
            & 4K 
            & 8K 
            & 16K 
            \\
            \midrule
            MambaInLlama \citep{mambainllama}&-&-&-&-&-&-&-&-&-&38.75& 21.55& 3.88\\
            Zebra-Llama \citep{zebrallama} &-&-&-&-&-&-&-&-&-&35.75& 24.80& 13.37\\
            SWA  &9.4  & 8.2 & 9.0 & 16.4 & 10.2 & 10.6 &11.4 & 4.4 & 7.2 & 6.00 &3.96 & 4.66\\
            LoLCATs~\citep{lolcats}  & 84.8 & 81.6 &  36.2 & 86.8 & 33.0 & 8.6 & 54.6 & 16.8 & 3.4 &
            33.99 & 17.93 & 7.59 \\
            \rowcolor[HTML]{E2ECF9}  
            STILL (Ours) & \textbf{100} & \textbf{100} & \textbf{100} & \textbf{100} & \textbf{99.0} & \textbf{74.0} & \textbf{70.4} & \textbf{67.2} & \textbf{25.8} & \textbf{53.24} &\textbf{35.66} &\textbf{27.20} \\
            \bottomrule
        \end{tabular}
    }
    \end{small}
    \vspace{-1ex}
\end{table}

\subsection{Ablation of Token-Routing Metrics}
\label{sec:routing_ablation}

Because Self-Saliency is related to diagonal self-attention, we ask whether diagonal attention alone is sufficient for routing. We therefore compare it with simpler selection criteria under the same routing and cache settings, covering both short-context reasoning and long-context retrieval.

\paragraph{Short-context reasoning.}
Table~\ref{tab:routing_short} reports the average performance over PIQA, ARC-E, ARC-C, HellaSwag, WinoGrande, and MMLU. Replacing Self-Saliency with the diagonal self-attention weight reduces the average score only slightly by 0.5\% This modest gap is expected because these benchmarks contain relatively short contexts, where retaining sparse long-range evidence is not the primary bottleneck. Thus, diagonal attention already captures much of the dominant routing trend.


\begin{table}[t]
    \centering
    \begin{minipage}[t]{0.43\textwidth}
        \centering
        \small
        \caption{Routing-metric ablation on short-context reasoning. $\Delta$ denotes the difference from Self-Saliency.}
        \label{tab:routing_short}
        \vspace{2pt}
        \setlength{\tabcolsep}{5pt}
        \resizebox{\linewidth}{!}{
        \begin{tabular}{lcc}
            \toprule
            Routing Metric & Avg. $\uparrow$ & $\Delta$ \\
            \midrule
            \rowcolor[HTML]{E2ECF9}
            Self-Saliency      & \textbf{72.5} & -- \\
            Diagonal attention & 72.0          & -0.5 \\
            \bottomrule
        \end{tabular}
        }
    \end{minipage}
    \hfill
    \begin{minipage}[t]{0.54\textwidth}
        \centering
        \small
        \caption{Routing-metric ablation on long-context S-NIAH-1 retrieval under the same 128 cache budget.}
        \label{tab:routing_long}
        \vspace{2pt}
        \setlength{\tabcolsep}{4pt}
        \resizebox{\linewidth}{!}{
        \begin{tabular}{l|llc}
            \toprule
            Routing Metric & 0.5K $\uparrow$& 1K $\uparrow$ & Avg. $\uparrow$ \\
            \midrule
            \rowcolor[HTML]{E2ECF9}
            Self-Saliency      & \textbf{63.2}    & \textbf{22.0}  & \textbf{42.6} \\
            Diagonal attention & $35.6_{\textcolor{red}{-27.6}}$ & $10.6_{\textcolor{red}{-11.4}}$ & 23.1 \\
            $\mathbf{q_t^\top k_t}$     & $35.4_{\textcolor{red}{-27.8}}$ & $10.2_{\textcolor{red}{-11.8}}$ & 22.8 \\
            Key norm           & $18.6_{\textcolor{red}{-44.6}}$ & $7.1_{\textcolor{red}{-14.9}}$ & 12.9 \\
            Uniform per chunk  & $17.8_{\textcolor{red}{-45.4}}$ & $6.8_{\textcolor{red}{-15.2}}$ & 12.3 \\
            Random selection   & $18.6_{\textcolor{red}{-44.6}}$ & $7.0_{\textcolor{red}{-15.0}}$ & 12.8 \\
            \bottomrule
        \end{tabular}
        }
    \end{minipage}
    \vspace{-3ex}
\end{table}

\paragraph{Long-context retrieval.}
The difference becomes substantially larger when routing is the key bottleneck. Table~\ref{tab:routing_long} compares Self-Saliency with diagonal attention, the query--key inner product, key norm, uniform per-chunk selection, and random selection on S-NIAH-1 under the same strict 128-token cache budget. Self-Saliency achieves 63.2 and 22.0 accuracy at 0.5K and 1K contexts, respectively, yielding an average of 42.6. In comparison, diagonal attention and $q_t^\top k_t$ obtain averages of only 23.1 and 22.8, while key-norm, uniform, and random routing remain around 12--13. These results show that although diagonal attention provides a reasonable proxy in ordinary short-context settings, it is not sufficient when accurate sparse retention of remote evidence is required. Self-Saliency is therefore particularly beneficial in the routing-sensitive long-context regime.


\subsection{Efficiency Comparison}
To assess the efficiency of STILL, we measure peak GPU memory usage and end-to-end latency on a single NVIDIA H200 GPU, varying the sequence length from 1K to 64K.
Results demonstrate that STILL achieves a favorable memory--speed trade-off.



\noindent\textbf{Prefilling.} 
Fig.~\ref{fig:efficiency-prefill} illustrates prefilling efficiency. Softmax attention hits memory bottlenecks beyond 16K tokens, whereas STILL scales smoothly to 64K, requiring only 17\% of the Softmax runtime at 16K.  Notably, our delayed selection strategy ensures STILL adds only marginal latency and memory overheads over LoLCATs, a minimal trade-off for its massive recovery of long-context accuracy.


\noindent\textbf{Decoding.}
Fig.~\ref{fig:efficiency-decoding} shows decoding efficiency. While Softmax scales poorly and OOMs beyond 32K tokens, STILL and LoLCATs maintain near-constant memory up to 64K, consuming a mere 20\% of the Softmax memory at 16K. Although STILL exhibits slightly higher latency than LoLCATs, it achieves stable throughput with a flat memory profile, proving its practicality for extended generation.

\begin{table}[h]
\vspace{-1ex}
    \centering
    \begin{minipage}[h]{0.49\linewidth}
        \centering
        \centerline{\includegraphics[width=\columnwidth]{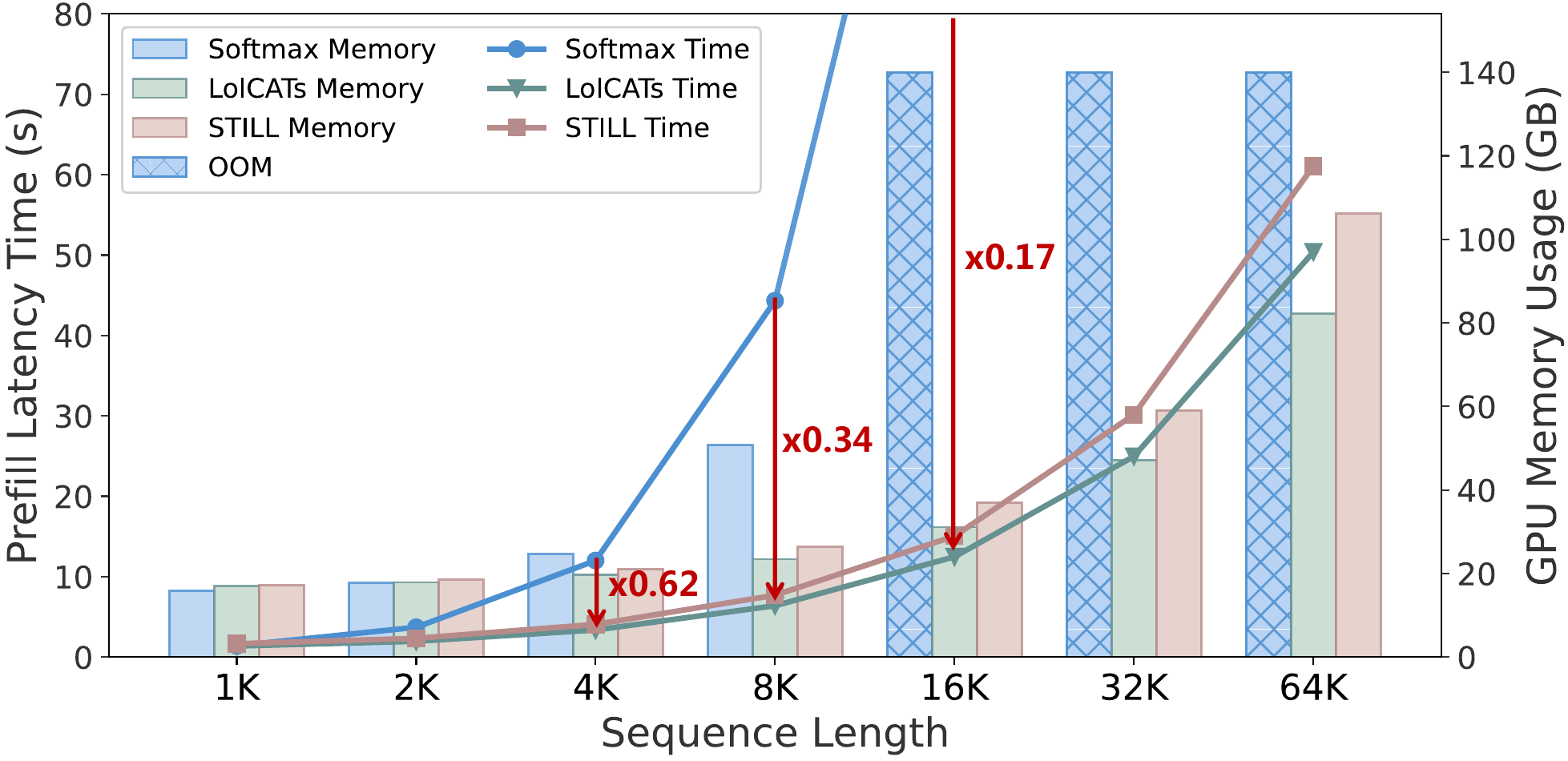}}
        \captionof{figure}{\textbf{Prefilling Efficiency.} STILL demonstrates favorable prefill latency and memory consumption from 1K to 64K.}
        \label{fig:efficiency-prefill}
    \end{minipage}
    \hfill
    \begin{minipage}[h]{0.49\linewidth}
        \centering
        \centerline{\includegraphics[width=\columnwidth]{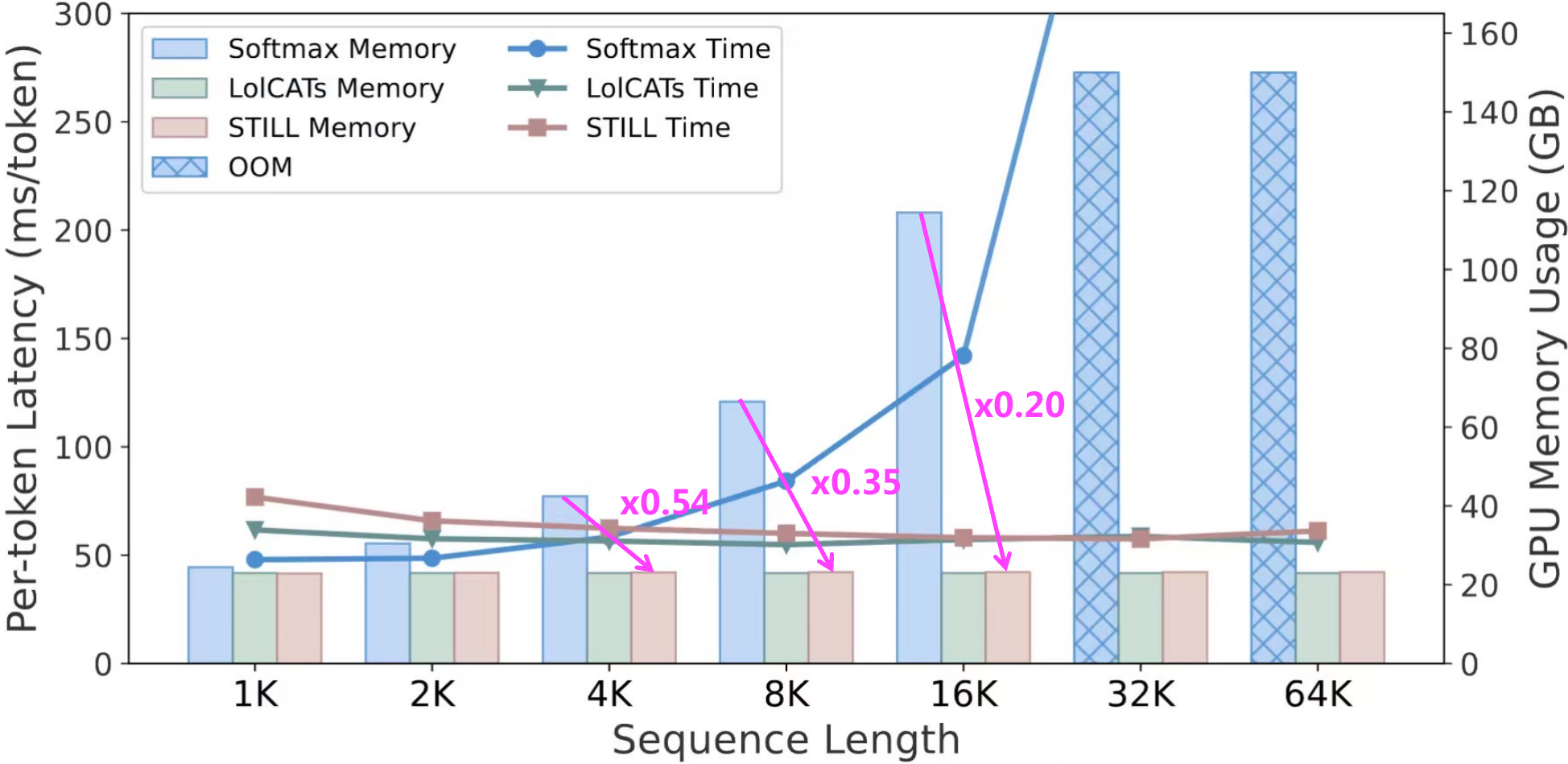}}
        \captionof{figure}{\textbf{Decoding Efficiency.} STILL maintains constant GPU memory while achieving lower latency for long-context decoding.}
        \label{fig:efficiency-decoding}
    \end{minipage}
    \vspace{-3ex}
\end{table}



\vspace{1ex}
\section{Conclusion}
\vspace{-1ex}
In this paper, we propose STILL, an intra-layer hybrid attention framework for efficiently linearizing pretrained LLMs. 
STILL leverages a self-saliency score to route the context tokens into softmax attention and linear attention. 
To preserve pretrained representational geometry, we introduce a norm-preserved learnable feature map that explicitly maintains norm consistency during linearization, and adopt a unified chunk-wise and delayed selection formulation to improve efficiency and parallelism. 
Together, these designs enable linear-time attention while largely retaining the reasoning and long-context capabilities of the original pretrained models.
\textcolor{black}{Through comprehensive evaluations across both intra-layer and intra-inter hybrid architectures, we demonstrate that STILL not only recovers long-context capabilities and improves short-sequence performance, but also delivers substantial memory savings and acceleration for both prefill and decoding.}


\bibliographystyle{plainnat}
\setlength{\bibhang}{0pt}
\setlength\bibindent{0pt}
\bibliography{main}

\newpage
\appendix

\newpage
\appendix
\onecolumn
\section{Appendix}

\begin{itemize}
    \item \ref{appendix:limitations} \textbf{Limitations.}
    \item \ref{appendix:related_works} \textbf{Related Work.}
    \item \ref{appendix:proof} \textbf{Theoretical Proof.} The proof of theorem.
    \item \ref{appendix:algorithm} \textbf{Algorithm.} Chunk-wise parallel form with delayed selection strategy.
    \item \ref{appendix:morevisualization} \textbf{More Visualizations.} More visualizations of Self-Saliency Score Distribution
    \item \ref{appendix:experimentalsetting} \textbf{Experimental Setups.}
    \item \ref{appendix:ablation} \textbf{Ablation Study.} Ablation of components and different backbone models and scales.
    \item \ref{appendix:tables} \textbf{Full Tables.} Full tables about the experimental results.
\end{itemize}

\subsection{Limitations}
\label{appendix:limitations}
While STILL demonstrates strong performance in linearizing LLMs, we acknowledge a few practical constraints. First, due to computational limits, our empirical evaluations focus on models up to the 8B parameter scale (e.g., Llama 3/3.1 8B, Mistral 7B). Although our theoretical derivations for the intra-layer hybrid routing are scale-agnostic, verifying transferability to massive models (e.g., 70B+) remains future work. Second, STILL is currently tailored for text-based autoregressive models; extending it to multimodal architectures is an open direction. Finally, while our fixed chunk and window configurations  yield highly competitive results, exploring dynamic, task-adaptive window sizing could further optimize hardware efficiency.

\subsection{Related Work}
\label{appendix:related_works}

\noindent\textbf{Linear Attention.}
As the central component of Transformer~\citep{attention}, self-attention effectively captures global dependencies among tokens.
In decoder-only LLMs~\citep{llama,mistral}, for each head, the current query $\mathbf{q}_t$ attends to all past key--value (KV) pairs, denoted as $\{\mathbf{k}_i\}_{i=1}^t \in \mathbb{R}^{t \times d}$ and $\{\mathbf{v}_i\}_{i=1}^t \in \mathbb{R}^{t \times d}$.
The attention scores are computed via dot product, normalized using softmax under causal masking, and used to aggregate the values. Formally, the output  $\mathbf{y}_t$ is given by 
\begin{equation}
    \mathbf{y}_{t} = \frac{ \sum_{i=1}^{t}\exp ( \mathbf{q}_t \mathbf{k}_i^\top / \sqrt{d})\mathbf{v}_i }
    { \sum_{j=1}^{t} \exp\ ( \mathbf{q}_t \mathbf{k}_j^\top / \sqrt{d}) }.
    \label{eq:softmax_attention} 
\end{equation}
While softmax attention provides a powerful mechanism to model long-range dependencies, its computation requires evaluating all pairwise interactions between queries and keys, resulting in $\mathcal{O}(N^2d)$ complexity. This quadratic cost becomes a significant bottleneck for both training and inference on long sequences.

To mitigate the efficiency bottlenecks of quadratic softmax attention, Linear Attention \citep{linear_attention} is proposed to approximate the softmax via a designed feature map $\phi(\cdot):\mathbb{R}^{d}\rightarrow\mathbb{R}^{d'}$ such that $\exp(\mathbf{q}\mathbf{k}^\top)\approx \phi(\mathbf{q})\phi(\mathbf{k})^\top$.
As shown in Equation~\eqref{eq:linear_attention}, by prioritizing the computation of the KV pairs, this approach reduces the complexity from $\mathcal{O}(N^2 d)$ to $\mathcal{O}(N {d'}^2)$.
\begin{equation}
    \mathbf{y}_t=\frac{\sum_{i=1}^{t}\phi(\mathbf{q}_t) \phi(\mathbf{k}_i)^\top\mathbf{v}_i}{\sum_{j=1}^{t}\phi(\mathbf{q}_t) \phi(\mathbf{k}_j)^\top}=\frac{\phi(\mathbf{q}_t)\sum_{i=1}^{t}\phi(\mathbf{k}_i)^\top\mathbf{v}_i}{\phi(\mathbf{q}_t)\sum_{j=1}^{t}\phi(\mathbf{k}_j)^\top}
    =\frac{\phi(\mathbf{q}_t)\mathbf{s}_t}{\phi(\mathbf{q}_t)\mathbf{z}_t}, 
    \label{eq:linear_attention}
\end{equation}
\begin{equation}
    \mathbf{s}_t=\mathbf{s}_{t-1}+\phi(\mathbf{k}_t)^\top\mathbf{v}_t,\quad\mathbf{z}_t=\mathbf{z}_{t-1}+\phi(\mathbf{k}_t)^\top.
    \label{eq:kvstate}
\end{equation}
The design of kernel functions generally considers two key factors: ensuring \textit{non-negative feature maps}, e.g., through $\operatorname{ReLU}(\cdot)$~\citep{cosformer,efficientvit} or $1 + \operatorname{ELU}(\cdot)$~\citep{linear_attention}, and promoting a \textit{spikiness property} to sharpen the attention distribution. Existing methods such as Flatten Transformer~\citep{flatten}, PolaFormer~\citep{polaformer,polaformer++}, and NaLaFormer~\citep{nalaformer} address the latter by designing power-based feature maps that produce low-entropy, softmax-like attention weights, enhancing the selectivity of attention. In addition, Hedgehog~\citep{hedgehog} uses a learnable feature map trained with a loss to approximate the exponential function, producing attention distributions closer to softmax.

Beyond the aforementioned designs targeting attention similarity, in autoregressive architectures such as LLMs, linear attention can be reformulated as a recurrent computation over the KV state as $\mathbf{s}_t = \mathbf{k}_t^\top \mathbf{v}_t$, making the state update strategy crucial (shown in Equation~\eqref{eq:kvstate}). RetNet~\citep{retnet} and TransnormerLLM~\citep{transnormerllm} use a decay factor to mitigate attention dilution. GLA~\citep{gla} introduces a gating mechanism on the KV state to selectively retain or forget information, while DeltaNet~\citep{deltanet} leverages a delta-rule update to dynamically adjust key--value associations for more effective long-context modeling. However, these methods cannot reuse pre-trained Transformer weights, requiring costly retraining from scratch and limiting the use of existing large-scale models.

\noindent\textbf{Linearized Transformer.}
Recently, several studies have explored leveraging existing pre-trained large language models (LLMs) to obtain linearized LLMs through low-cost training approaches.
These methods~\citep{t2r,supra,llamba} inherit pre-trained weights and linearize quadratic LLMs through uptraining, distillation, or fine-tuning.
However, they still require large token budgets and high memory during adaptation.
Consequently, some studies focus on achieving Transformer linearization with lower memory usage and fewer tokens.
LoLCATs~\citep{lolcats} linearizes pre-trained LLMs using only $40$M tokens via kernel distillation and low-rank fine-tuning.
Liger~\citep{liger} further reuses the key projection weights to construct gating, avoiding the introduction of additional parameters, and achieves linearization with only low-rank fine-tuning.
While these methods employ sliding-window attention to preserve local information sensitivity, they rely exclusively on linear attention for long-range modeling. This reliance significantly impairs the representation of long-range dependencies, leading to a pronounced degradation in performance on long-sequence tasks.
Therefore, a method is needed to properly allocate dependencies between softmax attention and linear attention, thereby restoring performance on long-sequence tasks.

\noindent\textbf{Token Selection and Sparse Attention Inference.}
Another line of research improves long-context inference by selectively retaining or retrieving salient tokens from the KV cache of pretrained softmax-attention models. Representative methods exploit accumulated attention statistics \citep{H2O}, observation-window attention and clustering \citep{snapkv}, query-aware block retrieval \citep{quest}, attention sinks \citep{streamllm}, layer-wise budget allocation \citep{Pyramidkv}, or head-level functional specialization \citep{duoattn}; related work further sparsifies attention computation during prefilling \citep{minference}. These approaches primarily reduce the inference cost of an otherwise unchanged quadratic attention model by pruning, compressing, or selectively accessing its KV cache. In contrast, STILL targets \emph{model linearization}: it converts pretrained softmax attention into a trainable intra-layer hybrid operator, where salient tokens retain high-fidelity softmax attention while the remaining context is continuously summarized by a recurrent linear-attention state. Moreover, Self-Saliency is computed causally from local attention, making the same routing mechanism applicable to both prefilling and autoregressive decoding.

\noindent\textbf{Recent Hybrid Distillation Advances.}
Beyond intra-layer designs, recent work improves Transformer-to-hybrid distillation along
orthogonal axes: layer-level softmax budget allocation
\citep{halo}, post-hoc architecture search over converted blocks
\citep{jetnemotron}, and rapid attention distillation pipelines
\citep{radalas}. Concurrent to our work, NAtS-L \citep{natsl}
routes chunks between softmax and gated delta-rule attention. STILL is complementary to these advances: our self-saliency routing and norm-preserved feature map operate \emph{within} each converted layer and can be composed with layer-selection strategies.


\clearpage

\subsection{Theoretical Proof}
\label{appendix:proof}
In this section, we provide the theoretical derivation demonstrating that the Self-Saliency Score effectively tracks the global self-attention probability and exhibits strong local-global consistency. This justifies our strategy of using local sliding-window computations to efficiently route tokens for global attention.

\begin{lemma}[Monotonicity of the Saliency Score]
Given an infinitesimally small smoothing parameter $\epsilon > 0$, the Self-Saliency Score $\operatorname{Score}_t$ is a strictly monotonically increasing function of the local self-attention probability $p_{local}$.
\end{lemma}

\begin{proof}
The Self-Saliency Score measures the divergence between the local attention distributions with and without the current token's self-similarity. Let $Z_t$ be the denominator of the softmax function for the local window, and $Z_{t \setminus t}$ be the denominator excluding the self-interaction term, such that:
\begin{equation}
Z_t = Z_{t \setminus t} + \exp(q_t k_t^\top)
\end{equation}
By definition, the local self-attention probability is $p_{local} = a_{t,t}^{diag} = \frac{\exp(q_t k_t^\top)}{Z_t}$. The total probability mass allocated to the other context tokens is:
\begin{equation}
1 - p_{local} = \frac{Z_{t \setminus t}}{Z_t}
\end{equation}
For any other context token ($j \neq t$), we can express its attention probability including the self-interaction term ($a_{t,j}^{diag}$) in terms of the probability without it ($a_{t,j}^{nodiag}$):
\begin{equation}
a_{t,j}^{diag} = \left( \frac{\exp(q_t k_j^\top)}{Z_{t \setminus t}} \right) \cdot \left( \frac{Z_{t \setminus t}}{Z_t} \right) = a_{t,j}^{nodiag} \cdot (1 - p_{local})
\end{equation}
Substituting this into the formulation of the Self-Saliency Score, and omitting the negligible $\epsilon$ in the non-diagonal terms for analytical simplicity as $\epsilon \to 0^+$, we obtain:
\begin{equation}
\operatorname{Score}_t \approx p_{local} \log \frac{p_{local}}{\epsilon} + \sum_{j \in \mathcal{W}_t \setminus \{t\}} (1 - p_{local}) a_{t,j}^{nodiag} \log(1 - p_{local})
\end{equation}
Since the probabilities sum to one ($\sum_{j \in \mathcal{W}_t \setminus \{t\}} a_{t,j}^{nodiag} = 1$), this simplifies to:
\begin{equation}
\operatorname{Score}_t \approx p_{local} \log \frac{p_{local}}{\epsilon} + (1 - p_{local}) \log(1 - p_{local})
\end{equation}
Taking the derivative with respect to $p_{local}$ yields:
\begin{equation}
\frac{\partial \operatorname{Score}_t}{\partial p_{local}} = \log \frac{p_{local}}{\epsilon} + 1 - \log(1 - p_{local}) - 1 = \log \frac{p_{local}}{\epsilon(1 - p_{local})}
\end{equation}
For an infinitesimally small $\epsilon > 0$ and any valid $p_{local} > 0$, the ratio $\frac{p_{local}}{\epsilon(1 - p_{local})} > 1$. Therefore, the derivative is strictly positive ($\frac{\partial \operatorname{Score}_t}{\partial p_{local}} > 0$), proving that $\operatorname{Score}_t$ increases strictly monotonically with $p_{local}$.
\end{proof}

\textbf{Remark:} Practical applicability condition.
\label{rem:eps-gate}
Lemma 1 is established for $\epsilon \to 0^{+}$. With a finite $\epsilon$,
the monotone relationship between $\mathrm{Score}_t$ and $p_{\mathrm{local}}$
is conditional: when $p_{\mathrm{local}} \gg \epsilon$, the two quantities
follow a similar trend and $\epsilon$ is negligible; when
$p_{\mathrm{local}}$ is comparable to $\epsilon$, the $\epsilon$-smoothed
terms dominate, and tokens are re-ranked according to the complete
leave-one-out distribution rather than the diagonal term alone. In this
regime $\epsilon$ acts as a soft gate: a token with a smaller
$a^{\mathrm{diag}}_{t,t}$ can still obtain a high score if removing its
self-interaction induces a larger change to the local attention
distribution. This is precisely the regime in which Self-Saliency departs
from --- and empirically outperforms --- the diagonal-attention baseline
(Table~\ref{tab:routing_long}).

\begin{lemma}[Probabilistic Local-Global Consistency]

Assuming that the attention logits for context tokens (i.e., $\exp(q_t k_j^\top)$ for $j \neq t$) are independent and identically distributed (i.i.d.) samples from a context distribution with expectation $C$, the global self-attention probability $p_{global}$ is a strictly monotonically increasing function of the local self-attention probability $p_{local}$.
\end{lemma}

\begin{proof}
Let $w$ be the local window size, $N$ be the global sequence length, and $S_t = \exp(q_t k_t^\top)$. By the Law of Large Numbers, for sufficiently large sequence lengths, the local and global partition functions (excluding the query token itself) converge to the expectation $C$:
\begin{equation}
\frac{1}{w} Z_{local} \xrightarrow{p} C, \quad \frac{1}{N} Z_{global} \xrightarrow{p} C \implies Z_{global} \approx \frac{N}{w} Z_{local}
\end{equation}
From the definition of $p_{local} = \frac{S_t}{S_t + Z_{local}}$, we can express the local partition function as $Z_{local} = S_t \frac{1 - p_{local}}{p_{local}}$.

The global self-attention probability is defined as $p_{global} = \frac{S_t}{S_t + Z_{global}}$. Substituting $Z_{global} = \frac{N}{w} Z_{local}$ into this formula, we obtain a direct analytical mapping between the local and global probabilities:
\begin{equation}
p_{global} = \frac{S_t}{S_t + \frac{N}{w} S_t \frac{1 - p_{local}}{p_{local}}} = \frac{1}{1 + \frac{N}{w} \left( \frac{1}{p_{local}} - 1 \right)} := g(p_{local})
\end{equation}
Letting $\gamma = \frac{N}{w}$ (where $\gamma \ge 1$), we compute the derivative of the mapping function $g(x)$:
\begin{equation}
g'(x) = \frac{\gamma}{x^2 \left( 1 + \gamma \left( \frac{1}{x} - 1 \right) \right)^2}
\end{equation}
For any valid probability $x \in (0, 1)$ and $\gamma > 0$, the derivative $g'(x) > 0$ strictly holds.

Since $g'(x) > 0$, the global self-attention probability is, in expectation, a strictly monotonic function of its local counterpart. Therefore, under the stated assumptions, the local Self-Saliency ranking provides a rank-preserving proxy for its global counterpart.
\end{proof}

\clearpage
\subsection{Algorithm}
\label{appendix:algorithm}

\begin{algorithm}[!h]
\caption{Chunk-Wise Parallel Form}
\label{alg:code}
\begin{algorithmic}

    \State \textbf{Input:}
    $\mathbf{q},\mathbf{k},\mathbf{v}\in\mathbb{R}^{N\times d}$,
    gate $\mathbf{g}\in\mathbb{R}^{N\times d}$,
    feature maps $\phi_q,\phi_k$,
    total cache budget $m$,
    local-window budget $W$,
    configured maximum length $N_{\max}$,
    chunk size $C$, and small constant $\epsilon$.

    \State $T_{\max}\leftarrow\lceil N_{\max}/C\rceil$
    \State $\lambda\leftarrow\lfloor(m-W)/T_{\max}\rfloor$
    \State $T\leftarrow\lceil N/C\rceil$
    \Comment{$T\lambda\le m-W$}

    \State Divide $\mathbf{q},\mathbf{k},\mathbf{v},\mathbf{g}$ into
    $T$ chunks of size $C\times d$.

    \State Initialize masks
    $\mathbf{M}^{\mathrm{down}},
    \mathbf{M}^{\mathrm{up}},
    \mathbf{M}^{\mathrm{diag}}
    \in\{0,1\}^{C\times C}$, where
    $\mathbf{M}^{\mathrm{down}}_{t,s}=1$ if $t\ge s$ else $0$,
    $\mathbf{M}^{\mathrm{up}}_{t,s}=1$ if $t<s$ else $0$, and
    $\mathbf{M}^{\mathrm{diag}}_{t,s}=1$ if $t>s$ else $0$.

    \State Define
    $\operatorname{Mask}(\mathbf{X},\mathbf{M})_{i,j}
    =
    \begin{cases}
    \mathbf{X}_{i,j}, & \mathbf{M}_{i,j}=1,\\
    -\infty, & \mathbf{M}_{i,j}=0.
    \end{cases}$

    \State \textbf{Compute Self-Saliency Score:}

    \State\quad
    $\mathbf{s}^{\mathrm{in}}
    \leftarrow
    \mathbf{q}\mathbf{k}^{\top}
    \in\mathbb{R}^{T\times C\times C}$.

    \State\quad
    $\mathbf{k}^{\rightarrow}_{[t,:,:]}
    \leftarrow
    \mathbf{k}_{[t-1,:,:]}$
    for $t>1$, and zeros for $t=1$.

    \State\quad
    $\mathbf{s}^{\mathrm{prev}}
    \leftarrow
    \mathbf{q}(\mathbf{k}^{\rightarrow})^{\top}
    \in\mathbb{R}^{T\times C\times C}$.

    \State\quad
    $\mathbf{a}^{\mathrm{diag}}
    \leftarrow
    \mathrm{softmax}\left(
    \left[
    \operatorname{Mask}
    (\mathbf{s}^{\mathrm{in}},\mathbf{M}^{\mathrm{down}}),
    \operatorname{Mask}
    (\mathbf{s}^{\mathrm{prev}},\mathbf{M}^{\mathrm{up}})
    \right]
    \right)$.

    \State\quad
    $\mathbf{a}^{\mathrm{nodiag}}
    \leftarrow
    \mathrm{softmax}\left(
    \left[
    \operatorname{Mask}
    (\mathbf{s}^{\mathrm{in}},\mathbf{M}^{\mathrm{diag}}),
    \operatorname{Mask}
    (\mathbf{s}^{\mathrm{prev}},\mathbf{M}^{\mathrm{up}})
    \right]
    \right)$.

    \State\quad
    $\mathbf{S}
    \leftarrow
    \mathrm{sum}\left(
    \mathbf{a}^{\mathrm{diag}}
    \log
    \frac{
    \mathbf{a}^{\mathrm{diag}}+\epsilon
    }{
    \mathbf{a}^{\mathrm{nodiag}}+\epsilon
    },
    \mathrm{dim}=-1
    \right)
    \in\mathbb{R}^{T\times C}$.

    \State \textbf{Local-Global Consistent Selection:}

    \State\quad
    $\mathbf{M}^{\mathcal{SA}}
    \leftarrow
    \mathrm{Top}(\mathbf{S},\lambda)$,
    \quad
    $\mathbf{M}^{\mathcal{LA}}
    \leftarrow
    \overline{\mathbf{M}^{\mathcal{SA}}}$.

    \State\quad
    $\mathbf{k}^{\mathcal{SA}}
    \leftarrow
    \mathrm{Index}
    (\mathbf{k},\mathbf{M}^{\mathcal{SA}})
    \in\mathbb{R}^{T\times\lambda\times d}$,

    \State\quad
    $\mathbf{v}^{\mathcal{SA}}
    \leftarrow
    \mathrm{Index}
    (\mathbf{v},\mathbf{M}^{\mathcal{SA}})
    \in\mathbb{R}^{T\times\lambda\times d}$.

    \State\quad
    $\mathbf{k}^{\mathcal{LA}}
    \leftarrow
    \mathrm{Index}
    (\mathbf{k},\mathbf{M}^{\mathcal{LA}})
    \in\mathbb{R}^{T\times(C-\lambda)\times d}$,

    \State\quad
    $\mathbf{v}^{\mathcal{LA}}
    \leftarrow
    \mathrm{Index}
    (\mathbf{v},\mathbf{M}^{\mathcal{LA}})
    \in\mathbb{R}^{T\times(C-\lambda)\times d}$.

    \State \textbf{Linear Attention Branch:}

    \State\quad
    $\mathbf{s}^{\mathrm{chunk}}
    \leftarrow
    \phi_k(\mathbf{k}^{\mathcal{LA}})^{\top}
    \mathbf{v}^{\mathcal{LA}}
    \in\mathbb{R}^{T\times d'\times d}$.

    \State\quad
    $\mathbf{z}^{\mathrm{chunk}}
    \leftarrow
    \mathrm{sum}\left(
    \phi_k(\mathbf{k}^{\mathcal{LA}}),
    \mathrm{dim}=1
    \right)
    \in\mathbb{R}^{T\times d'}$.

    \State\quad
    $\mathbf{s}
    \leftarrow
    \mathrm{Cumsum}
    (\mathbf{s}^{\mathrm{chunk}},\mathrm{dim}=1)
    \in\mathbb{R}^{T\times d'\times d}$.

    \State\quad
    $\mathbf{z}
    \leftarrow
    \mathrm{Cumsum}
    (\mathbf{z}^{\mathrm{chunk}},\mathrm{dim}=1)
    \in\mathbb{R}^{T\times d'}$.

    \State\quad
    $\mathbf{s}^{\rightarrow}_{[t,:,:]}
    \leftarrow
    \mathbf{s}_{[t-2,:,:]}$
    and
    $\mathbf{z}^{\rightarrow}_{[t,:]}
    \leftarrow
    \mathbf{z}_{[t-2,:]}$
    for $t>2$, and zeros otherwise.

    \State \textbf{Softmax Attention Branch:}

    \State\quad
    $\mathbf{k}^{\mathcal{SA}}_{\mathrm{hist}}
    \leftarrow
    \mathrm{Broadcast}\left(
    \mathrm{Flatten}(\mathbf{k}^{\mathcal{SA}})
    \right)
    \in\mathbb{R}^{T\times(T\lambda)\times d}$.

    \State\quad
    $\mathbf{v}^{\mathcal{SA}}_{\mathrm{hist}}
    \leftarrow
    \mathrm{Broadcast}\left(
    \mathrm{Flatten}(\mathbf{v}^{\mathcal{SA}})
    \right)
    \in\mathbb{R}^{T\times(T\lambda)\times d}$.

    \State\quad
    Initialize mask
    $\mathbf{M}^{\mathrm{Chunk}}
    \leftarrow
    \mathbf{0}
    \in\{0,1\}^{T\times(T\lambda)}$.

    \State\quad
    $\mathbf{M}^{\mathrm{Chunk}}_{[t,\;1:\max(0,t-2)\lambda]}
    \leftarrow1,
    \quad
    \forall t\in\{1,\ldots,T\}$.

    \State\quad
    $\mathbf{s}^{\mathcal{SA}}_{\mathrm{hist}}
    \leftarrow
    \operatorname{Mask}\left(
    \mathbf{q}
    (\mathbf{k}^{\mathcal{SA}}_{\mathrm{hist}})^{\top},
    \mathbf{M}^{\mathrm{Chunk}}
    \right)$.

    \State\quad
    $\mathbf{o}^{\mathcal{SA}}
    \leftarrow
    \exp\left(
    \left[
    \mathbf{s}^{\mathcal{SA}}_{\mathrm{hist}},
    \operatorname{Mask}
    (\mathbf{s}^{\mathrm{prev}},\mathbf{M}^{\mathrm{up}}),
    \operatorname{Mask}
    (\mathbf{s}^{\mathrm{in}},\mathbf{M}^{\mathrm{down}})
    \right]
    \right)$.

    \State\quad
    $\mathbf{v}^{\rightarrow}_{[t,:,:]}
    \leftarrow
    \mathbf{v}_{[t-1,:,:]}$
    for $t>1$, and zeros for $t=1$.

    \State\quad
    $\widetilde{\mathbf{v}}^{\mathcal{SA}}
    \leftarrow
    \left[
    \mathbf{v}^{\mathcal{SA}}_{\mathrm{hist}},
    \mathbf{v}^{\rightarrow},
    \mathbf{v}
    \right]$.

    \State\quad
    $\mathbf{y}
    \leftarrow
    \frac{
    \mathbf{o}^{\mathcal{SA}}
    \widetilde{\mathbf{v}}^{\mathcal{SA}}
    +
    \phi_q(\mathbf{q})
    \mathbf{s}^{\rightarrow}
    \odot\mathbf{g}
    }{
    \mathrm{sum}
    (\mathbf{o}^{\mathcal{SA}},\mathrm{dim}=-1)
    +
    \phi_q(\mathbf{q})
    \mathbf{z}^{\rightarrow}
    }
    \in\mathbb{R}^{T\times C\times d}$.

    \State \textbf{Output:}
    $\mathbf{y}
    \leftarrow
    \mathrm{Flatten}(\mathbf{y})
    \in\mathbb{R}^{N\times d}$.

\end{algorithmic}
\end{algorithm}

\vspace{-4ex}
\paragraph{Budget certification and runtime lengths.}
The per-chunk quota $\lambda$ is derived from the configured $N_{\max}$ so that the hard
budget $m$ is certified for every runtime length $N \le N_{\max}$
($T\lambda \le T_{\max}\lambda \le m-W$). For $N \ll N_{\max}$ the salient cache is
intentionally under-filled, trading utilization for a length-independent memory footprint.
When the deployment length is known, $\lambda$ can equivalently be re-derived from the
actual chunk count at inference, $\lambda(N) = \lfloor (m-W)/\lceil N/C \rceil \rfloor$,
filling the cache with no architectural or training change; the same re-derivation keeps
the budget exact for $N > N_{\max}$, degrading gracefully in selection density rather than
exceeding memory. All experiments use the certified static $\lambda$ for reproducibility.

\clearpage
\subsection{More Visualizations}
\label{appendix:morevisualization}
\begin{figure}[h]
    \centering
    \includegraphics[width=1\linewidth]{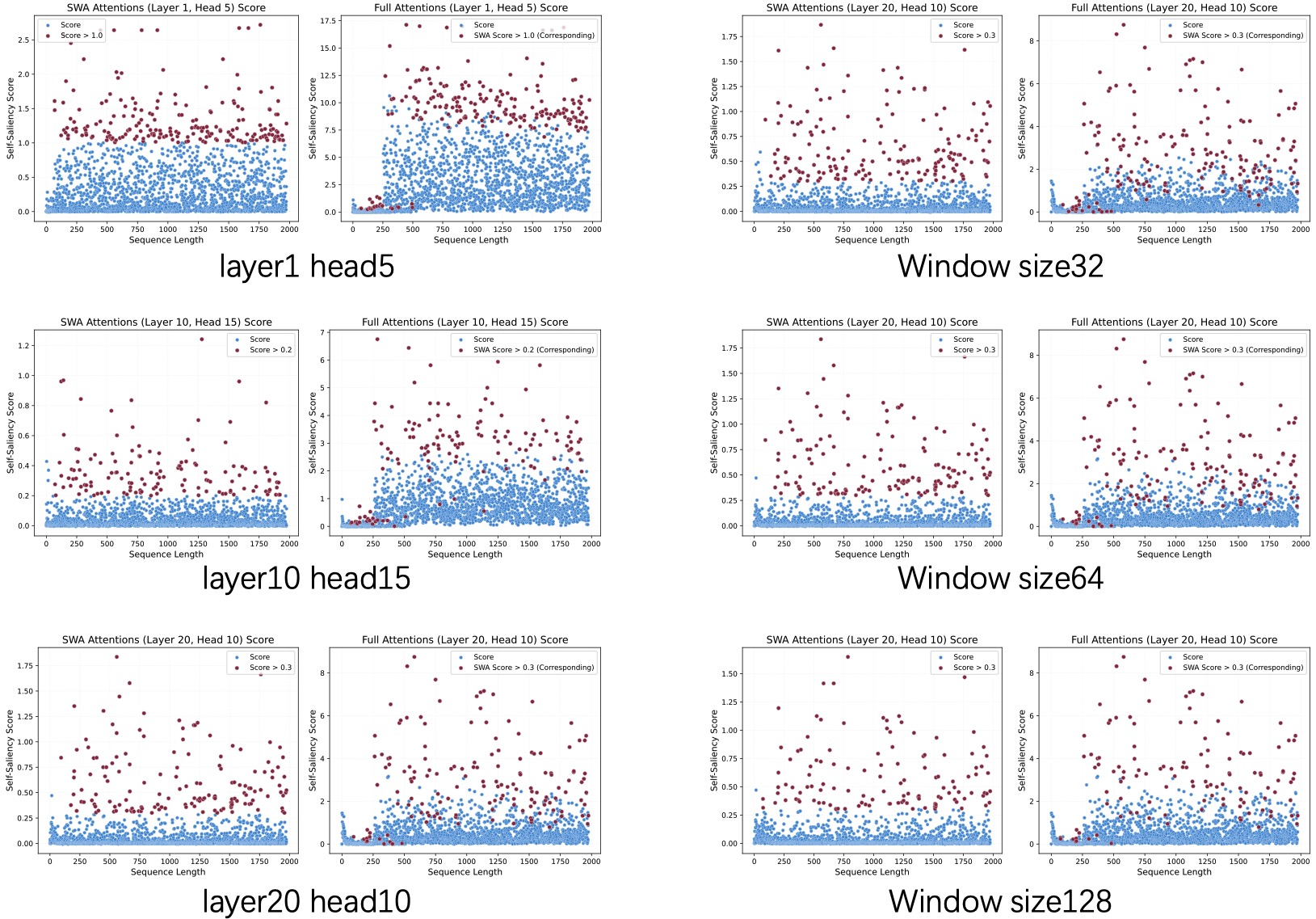}
    \caption{Self-saliency score comparison between local and global attention across different layers, heads, and window sizes.}
    \label{fig:placeholder}
\end{figure}

\subsection{Experimental Setups}
\label{appendix:experimentalsetting}



\subsubsection{Training Strategy and Datasets} 
To rigorously evaluate our proposed STILL module while strictly controlling computational variables against baselines, we introduce a primary lightweight training configuration, alongside an extended configuration to test architectural compatibility.

\textbf{Primary Configuration (Intra-Layer Validation).} For our core experiments (Section 4.2 and 4.3), we strictly adhere to the ultra-lightweight two-stage training procedure proposed in LoLCATs \citep{lolcats}. 
Specifically, during the \emph{Attention Transfer} stage, we freeze the teacher model parameters and optimize the feature mapping $\phi$ and the gating projection $\mathbf{W}_g$ by minimizing the layer-wise mean squared error (MSE) between the attention maps of the original and proposed modules. 
In the \emph{Low-rank Linearization} stage, we perform end-to-end training using Low-Rank Adaptation (LoRA) \citep{lora} applied to the query, key, value, and output projection matrices, as well as the gating mechanism ($\mathbf{W}_q, \mathbf{W}_k, \mathbf{W}_v, \mathbf{W}_o, \mathbf{W}_g$). 
Both stages are conducted on the Cleaned Alpaca dataset \citep{alpaca} utilizing a minimal budget of only 20M tokens with a constant sequence length of 1024.

\textbf{Extended Configuration (Architectural Compatibility).} To demonstrate STILL's plug-and-play flexibility within macroscopic layer-routing architectures (evaluated in Section 4.4), we integrate our intra-layer mechanism with the KL-LS inter-layer strategy~\citep{KLSelecting} on Llama 3.2 1B. Specifically, 25\% of the layers are retained as full softmax attention, while in the remaining layers, STILL selects 12.5\% of the tokens for local softmax attention. This dual-hybrid setup operates on the DCLM dataset~\citep{dclm} in two stages: an initial attention transfer phase (100M tokens, sequence length 512), followed by a full-parameter alignment step (250M tokens) adapted to a sequence length of 4096 to accommodate the structural changes introduced by the inter-layer routing.

We note that the routing score is length-invariant by construction: the Self-Saliency
Score is computed entirely within a local window of $W \le 512$ tokens (Eqs.~2--5), so its
distribution at long test contexts matches that at the 1024-token training sequences; what
is learned from the training corpus is the fidelity of the LA branch (NP-Map) and the LoRA
alignment, i.e., short-range behavior, while long-context capability emerges at inference
from the accumulation of per-chunk routing decisions. Our use of Cleaned Alpaca at length
1024 follows the established linearization protocol of LoLCATs \citep{lolcats} and
Liger \citep{liger}, ensuring identical training data and token budgets across all
compared methods in Secs.~4.2--4.3.

\subsubsection{Evaluation Settings}
\noindent\textbf{Commonsense and General Reasoning Tasks Settings.}
We evaluate the commonsense reasoning and general language understanding capabilities of STILL on a diverse suite of standard benchmarks, including PIQA~\citep{piqa}, ARC-Easy and ARC-Challenge~\citep{arc}, HellaSwag~\citep{hellaswag}, WinoGrande~\citep{winogrande}, and MMLU~\citep{mmlu}. Consistent with prior work~\citep{supra,lolcats,liger}, we evaluate all benchmarks in the zero-shot setting, with MMLU evaluated under the standard five-shot protocol. 
We select Llama 3 8B and Llama 3.1 8B~\citep{llamba} as backbone models, and compare our method against four categories of baselines: the original Transformer, subquadratic models trained from scratch, inter-layer hybrid models and linearized models.

\noindent\textbf{S-NIAH Settings.}
We evaluate the S-NIAH tasks~\citep{ruler} with context lengths up to 4K, which assess a model's ability to retrieve a target span from long inputs.
We compare STILL against sliding-window attention (SWA), LoLCATs and Liger-GLA with window sizes of 128, 256, and 512, using Llama 3.1 8B as the backbone model.
For STILL, we set the local-window size to half of the corresponding baseline configuration and match the overall cache budget by ensuring that the sum of cached tokens and the sliding-window size does not exceed the baseline window size.
We keep the training procedure of all models unchanged and only modify the sliding-window size during inference.

\noindent\textbf{Full RULER Settings.}
We further evaluate STILL on the extended RULER benchmark~\citep{ruler} at a 4K context length, which requires stronger memory and state tracking beyond NIAH-style retrieval. 
It includes multi-key retrieval (MK), multi-query retrieval (MQ), multi-value retrieval (MV), common word extraction (CWE), frequent word extraction (FWE), HotpotQA (HQA), SQuAD-QA (SQA), and variable tracking (VT). 
We compare STILL against Mamba2~\citep{mamba2}, LoLCATs~\citep{lolcats}, and Liger~\citep{liger}, using Llama 3.1 8B as the backbone model. LoLCATs and Liger are evaluated with a sliding-window size of 896. STILL is evaluated with a local-window size of 128, and the total number of selected tokens is capped at 896.

\noindent\textbf{BABILong Settings.}
To evaluate the long-context reasoning capability of STILL, we conduct experiments on BABILong \citep{babilong}, a long-context extension of the bAbI \citep{babi} question-answering tasks. BABILong embeds key supporting facts into very long irrelevant contexts, requiring models to integrate evidence across distant positions for reasoning rather than relying on simple keyword matching.
We compare STILL with SWA and LoLCATs under a total cache budget of 256, using Llama~3.1~8B as the backbone model.
Experiments are conducted across tasks QA1--QA5.

\clearpage
\subsection{Ablation Studies}
\label{appendix:ablation}

We conduct ablation studies to examine the robustness of STILL under different backbone architectures and model scales, and to quantify the impact of our key design components.

\subsubsection{Different Backbone Architectures and Scales.}
Tab.~\ref{tab:teacher} summarizes results across different backbone families and model scales. STILL consistently outperforms all linearized baselines in overall performance.
With the Llama 3.2 1B \citep{llama}, STILL improves the average score including MMLU to 54.8 and raises MMLU from 22.4 with Liger to 29.8. Using the stronger Llama 3.2 3B \citep{llama}, STILL further boosts MMLU to 52.4, compared to 32.1 for Liger, achieving an overall average of 65.5. For a linearized Mistral 7B \citep{mistral}, STILL attains an average score of 70.9 including MMLU, surpassing LoLCATs at 69.1 and Liger at 65.1.

\begin{table}[h]
    \caption{\textbf{Different Backbone Architectures and Scales Results.} STILL achieves the best performance across all models and scales.}
    \label{tab:teacher}
    \begin{center}
            \resizebox{1\linewidth}{!}{
                \begin{tabular}{l|c|ccccc|c|cc}
                    \toprule
                    \multirow{2}{*}{\textbf{\textsc{Model}}}
                    & \textbf{\textsc{Training}}
                    & \textbf{\textsc{PIQA}}
                    & \textbf{\textsc{ARC-e}}
                    & \textbf{\textsc{ARC-c}}
                    & \textbf{\textsc{Hella.}}
                    & \textbf{\textsc{Wino.}}
                    & \textbf{\textsc{MMLU}}
                    & \textbf{\textsc{Avg.}}
                    & \textbf{\textsc{Avg.}}
                    \\
                    & \textbf{\textsc{Tokens}} (B)
                    & acc $\uparrow$
                    & acc $\uparrow$ 
                    & acc$_n$ $\uparrow$
                    & acc$_n$ $\uparrow$  
                    & acc $\uparrow$  
                    & (5-shot)  $\uparrow$  
                    & (w. MMLU)
                    & (wo. MMLU)
                    \\
                    \midrule 
                    \rowcolor[HTML]{F1E9E7}
                    \multicolumn{10}{l}{\textbf{\textit{Linearized from Llama 3.2 1B}}}
                    \\
                    Llama 3.2 1B \citep{llama}
                    & 9000
                    & 74.4
                    & 65.5
                    & 35.8
                    & 63.7
                    & 60.5
                    & 31.9
                    & 55.3
                    & 60.0
                    \\
                    T2R \citep{t2r}
                    & 0.04
                    & 69.2
                    & 58.2
                    & 29.9
                    & 42.6
                    & 54.1
                    & \underline{23.3}
                    & 46.2
                    & 50.8
                    \\
                    Hedgehog \citep{hedgehog}
                    & 0.04
                    & 70.1
                    & 55.8
                    & 29.8
                    & 47.7
                    & 50.7
                    & 23.0
                    & 46.2
                    & 50.8
                    \\
                    LoLCATs \citep{lolcats}
                    & 0.04
                    & 74.1
                    & 63.7
                    & \textbf{36.4}
                    & 51.2
                    & 58.2
                    & 23.1
                    & 51.1
                    & 56.7
                    \\ 
                    Liger-GLA \citep{liger}
                    & 0.02
                    & \textbf{75.0}
                    & \textbf{65.4}
                    & 35.7
                    & \underline{59.8}
                    & \underline{59.1}
                    & 22.4
                    & \underline{52.9}
                    & \underline{59.0}
                    \\
                    \rowcolor[HTML]{F1F5FC} 
                    STILL
                    & 0.04
                    & \underline{74.9}
                    & \underline{65.3}
                    & \underline{36.3}
                    & \textbf{62.6}
                    & \textbf{59.8}
                    & \textbf{29.8}
                    & \textbf{54.8}
                    & \textbf{59.8}
                    \\
                    \midrule
                    \rowcolor[HTML]{F1E9E7}
                    \multicolumn{10}{l}{\textbf{\textit{Linearized from Llama 3.2 3B}}}
                    \\
                    Llama 3.2 3B \citep{llama}
                    & 9000
                    & 76.4
                    & 74.7
                    & 46.0
                    & 73.6
                    & 69.9
                    & 56.2
                    & 66.1
                    & 68.1
                    \\
                    LoLCATs \citep{lolcats}
                    & 0.04
                    & 76.7
                    & 72.0
                    & 42.3
                    & 51.9
                    & \underline{66.9}
                    & 23.6
                    & 55.6
                    & 62.0
                    \\ 
                    Liger-GLA \citep{liger}
                    & 0.02
                    & \textbf{77.9}
                    & \underline{74.0}
                    & \underline{43.9}
                    & \underline{70.3}
                    & 66.3
                    & \underline{32.1}
                    & \underline{60.7}
                    & \underline{66.5}
                    \\
                    \rowcolor[HTML]{F1F5FC} 
                    (Ours)
                    & 0.04
                    & \underline{77.6}
                    & \textbf{76.1}
                    & \textbf{47.5}
                    & \textbf{72.0}
                    & \textbf{67.6}
                    & \textbf{52.4}
                    & \textbf{65.5}
                    & \textbf{68.2}
                    \\ 
                    \midrule 
                    \rowcolor[HTML]{F1E9E7}
                    \multicolumn{10}{l}{\textbf{\textit{Linearized from Mistral 7B}}}
                    \\
                    Mistral 7B \citep{mistral}
                    & 8000
                    & 82.1
                    & 80.9
                    & 53.8
                    & 81.0
                    & 74.0
                    & 62.4
                    & 72.4
                    & 74.4
                    \\
                    SUPRA \citep{supra}
                    & 20
                    & 80.1
                    & 74.6
                    & 42.3
                    & 74.8
                    & 67.4
                    & 28.0
                    & 61.2
                    & 67.8
                    \\
                    SUPRA \citep{supra}
                    & 100
                    & 80.4
                    & 75.9
                    & 45.8
                    & 77.1
                    & 70.3
                    & 34.2
                    & 64.0
                    & 69.9
                    \\
                    LoLCATs \citep{lolcats}
                    & 0.04
                    & \underline{81.5}
                    & \underline{81.7}
                    & \underline{54.9}
                    & \underline{79.2}
                    & \underline{71.0}
                    & \underline{46.2}
                    & \underline{69.1}
                    & \underline{73.7}
                    \\ 
                    
                    Liger-GLA \citep{liger}
                    & 0.02
                    & 80.1
                    & 78.7
                    & 49.3
                    & 76.3
                    & 70.1
                    & 36.3
                    & 65.1
                    & 70.9
                    \\
                    \rowcolor[HTML]{F1F5FC} 
                    (Ours)
                    & 0.04
                    & \textbf{81.7}
                    & \textbf{82.7}
                    & \textbf{55.6}
                    & \textbf{80.0}
                    & \textbf{71.6}
                    & \textbf{53.8}
                    & \textbf{70.9}
                    & \textbf{74.3}
                    \\ 
 
                    \bottomrule
            \end{tabular}
        }
    \end{center}

\end{table}

\subsubsection{Module Ablation}
We further conduct a module ablation study of STILL on the BABILong benchmark with a sequence length of 1K, as shown in Tab.~\ref{tab:module-ablation}.
Removing either the gating mechanism or the non-parametric mapping (NP-Map) degrades performance by 5.8 and 4.9 points, respectively, and removing both causes an 8.3-point drop.
When the self-saliency-based token selection strategy is further removed, the accuracy drops from 27.5 to 12.5, demonstrating that each component contributes meaningfully and that they are complementary.

\begin{table}[ht]
    \centering
    \caption{\textbf{Module Ablation Results.} Ablation study of key components in STILL.}
    \label{tab:module-ablation}
    \setlength{\tabcolsep}{18pt}
        \begin{tabular}{ccc|l}
            \toprule
            \textbf{\textsc{Self-Saliency Score}} 
            & \textbf{\textsc{NP-Map}} 
            & \textbf{\textsc{Gate}}
            & \textbf{\textsc{Acc.}}
            \\
            \midrule
            $\checkmark$ & $\checkmark$ & $\checkmark$ &27.5 \\
            $\checkmark$ & $\checkmark$ &  & 21.7$_\text{\textcolor{red}{-5.8}}$ \\
            $\checkmark$ & & $\checkmark$ & 22.4$_\text{\textcolor{red}{-4.9}}$ \\
            $\checkmark$ &  &  & 19.2$_\text{\textcolor{red}{-8.3}}$ \\
             &  &  & 12.5$_\text{\textcolor{red}{-15.0}}$ \\
            \bottomrule
        \end{tabular}
\end{table}

\newpage
\subsubsection{Ablation of Token-Routing Metrics}
\label{appendix:routing_metric}

To examine whether the proposed Self-Saliency Score provides a more effective
routing signal than simpler alternatives, we compare it with several natural
token-selection criteria under the same routing and cache configuration.
We consider both short-context commonsense reasoning, where long-range cache
routing is relatively unimportant, and S-NIAH retrieval, where a small number
of relevant tokens must be retained among many distractors under a strict cache
budget.

As shown in Tab.~\ref{tab:routing_short_appendix}, replacing Self-Saliency with the
diagonal self-attention weight causes only a modest $0.5$-point degradation on
standard commonsense reasoning tasks. This is expected because these benchmarks
contain relatively short contexts, where retaining sparse long-range evidence
is not the primary bottleneck. In this regime, diagonal attention already
captures much of the dominant routing trend.

\begin{table}[ht]
    \centering
    \caption{
    \textbf{Routing-metric ablation on short-context commonsense reasoning.}
    We report the average performance over PIQA, ARC-E, ARC-C, HellaSwag,
    WinoGrande, and MMLU. $\Delta$ denotes the difference from Self-Saliency.
    }
    \label{tab:routing_short_appendix}
    \vspace{-0.5ex}
    \begin{tabular}{lcc}
        \toprule
        \textbf{Routing Metric} &
        \textbf{Avg. $\uparrow$} &
        \textbf{$\Delta$} \\
        \midrule
        \textbf{Self-Saliency}     & \textbf{72.5} & -- \\
        Diagonal attention         & 72.0          & -0.5 \\
        \bottomrule
    \end{tabular}
    \vspace{-1ex}
\end{table}

The distinction becomes substantially larger on S-NIAH
(Tab.~\ref{tab:routing_long_appendix}), where routing is central to preserving remote
evidence under a limited cache budget. Self-Saliency achieves an average
accuracy of $42.6$, compared with $23.1$ for diagonal attention and $22.8$ for
$\mathbf{q}_t^\top\mathbf{k}_t$, while norm-, uniform-, and random-based
selection perform considerably worse. These results indicate that diagonal
attention captures a strong local routing signal in ordinary short-context
settings, but does not provide an equally effective substitute when accurate
sparse long-range retention is required. Self-Saliency is particularly
beneficial in this routing-sensitive regime.

\begin{table}[ht]
    \centering
    \caption{
    \textbf{Routing-metric ablation on long-context S-NIAH retrieval.}
    All methods use the same cache budget. Avg. denotes the mean accuracy over
    the 0.5K and 1K context lengths, and $\Delta$ denotes the difference from
    Self-Saliency.
    }
    \label{tab:routing_long_appendix}
    \vspace{-0.5ex}
    \begin{tabular}{lccccc}
        \toprule
        \textbf{Routing Metric} &
        \textbf{0.5K $\uparrow$} &
        \textbf{$\Delta$} &
        \textbf{1K $\uparrow$} &
        \textbf{$\Delta$} &
        \textbf{Avg. $\uparrow$} \\
        \midrule
        \textbf{Self-Saliency}
            & \textbf{63.2} & --    & \textbf{22.0} & --    & \textbf{42.6} \\
        Diagonal attention
            & 35.6 & -27.6 & 10.6 & -11.4 & 23.1 \\
        $\mathbf{q}_t^\top\mathbf{k}_t$
            & 35.4 & -27.8 & 10.2 & -11.8 & 22.8 \\
        Key norm
            & 18.6 & -44.6 & 7.1  & -14.9 & 12.9 \\
        Uniform per chunk
            & 17.8 & -45.4 & 6.8  & -15.2 & 12.3 \\
        Random selection
            & 18.6 & -44.6 & 7.0  & -15.0 & 12.8 \\
        \bottomrule
    \end{tabular}
    \vspace{-1ex}
\end{table}

\subsubsection{Future KV utility}
Self-Saliency is computed causally within a token's local window, whereas
routing requires estimating whether the token will be attended by
\emph{future} queries after it becomes remote context. We directly measure
this correspondence. Under the standard delayed-selection procedure, we
select the Top-224 tokens with locally computed Self-Saliency, then run
full attention and rank the eligible remote keys --- excluding the
sliding-window and delayed-selection regions --- by the attention weights
assigned by subsequent queries. As shown in
Table~\ref{tab:future-kv-overlap}, the causal local score recovers
62.9\%--84.8\% of the remote keys actually prioritized by future
full-attention queries across representative heads in shallow (0), middle
(23, 24), and deep (28) layers. We do not include accumulated or
recent-received attention as a baseline: under delayed chunk routing it
requires an additional causal observation horizon and a normalization for
unequal exposure across positions, and maintaining such query-dependent
statistics contradicts the online, state-free nature of our routing. Oracle
selection with full attention is a non-deployable quadratic upper bound and
is instead reflected by the overlap measurement itself. Together with the
downstream retrieval results (Table~\ref{tab:routing_long_appendix}), these results
support Self-Saliency as an effective online proxy of future KV utility.

\begin{table}[ht]
\centering
\caption{Overlap between the Top-224 tokens selected by locally computed
Self-Saliency and the Top-224 remote keys prioritized by future
full-attention queries (Llama 3.1 8B).}
\label{tab:future-kv-overlap}
\begin{tabular}{ccccc}
\toprule
\textbf{Layer} & \textbf{Head} & \textbf{Selected} & \textbf{Overlap} & \textbf{Ratio} \\
\midrule
0  & 0 & 224 & 141 & 62.9\% \\
0  & 1 & 224 & 145 & 64.7\% \\
0  & 2 & 224 & 158 & 70.5\% \\
0  & 3 & 224 & 190 & 84.8\% \\
\midrule
23 & 0 & 224 & 185 & 82.6\% \\
23 & 1 & 224 & 188 & 83.9\% \\
23 & 2 & 224 & 188 & 83.9\% \\
23 & 3 & 224 & 143 & 63.8\% \\
\midrule
24 & 0 & 224 & 168 & 75.0\% \\
24 & 1 & 224 & 163 & 72.8\% \\
24 & 2 & 224 & 164 & 73.2\% \\
24 & 3 & 224 & 150 & 67.0\% \\
\midrule
28 & 0 & 224 & 145 & 64.7\% \\
28 & 1 & 224 & 180 & 80.4\% \\
28 & 2 & 224 & 169 & 75.4\% \\
28 & 3 & 224 & 150 & 67.0\% \\
\bottomrule
\end{tabular}
\vspace{-1em}
\end{table}

\subsubsection{Sensitivity to Chunk Size and Delayed Selection}
\label{app:chunk-sensitivity}

In STILL, the local window spans one chunk, so the chunk size $C$ jointly
controls the observation window of Self-Saliency and the routing delay. We
vary $C$ while keeping the total selection budget unchanged (16 selected
remote tokens) and using the same checkpoint.

\paragraph{Routing quality.}
Table~\ref{tab:chunk-acc} shows that delayed selection introduces no
systematic accuracy loss: performance improves from $C{=}16$ to $C{=}64$
and remains close at $C{=}128$, indicating that moderate chunk-level delay
preserves routing quality and can even outperform finer-grained selection
under the same budget. The lower performance at $C{=}16$ suggests that the
local window should not be too small, both to preserve short-range
generation quality and to support reliable saliency estimation.

\begin{table}[h]
\centering
\caption{Commonsense reasoning accuracy under different chunk sizes $C$
with a fixed selection budget.}
\label{tab:chunk-acc}
\resizebox{\linewidth}{!}{
\begin{tabular}{lcccccccc}
\toprule
$C$ & \textbf{PIQA} & \textbf{ARC-e} & \textbf{ARC-c} & \textbf{Hella.} & \textbf{Wino.} & \textbf{MMLU} & \textbf{Avg.\ w/ MMLU} & \textbf{Avg.\ w/o MMLU} \\
\midrule
16  & 80.5 & 81.3 & 53.4 & 77.2 & 71.0 & 54.2 & 69.6 & 72.7 \\
64  & 81.4 & 82.8 & 56.4 & 79.0 & 73.1 & 60.8 & 72.3 & 74.5 \\
128 & 81.3 & 82.7 & 56.2 & 79.0 & 71.1 & 62.1 & 72.1 & 74.1 \\
\bottomrule
\end{tabular}
}
\vspace{-1em}
\end{table}

\paragraph{Hardware efficiency.}
Table~\ref{tab:chunk-eff} reports training cost on a single NVIDIA A6000.
Increasing $C$ from 16 to 128 reduces training time by 35.4\% and memory by
21.7\%, while $C{=}4$ runs out of memory. Exact per-token routing
corresponds to $C{=}1$: it breaks chunk-level parallelism, and in separate
profiling it increases decoding selection latency by more than $3\times$
and reduces training speed by more than 50\%. Overall, $C{=}64$ provides
the best balance between routing quality and hardware efficiency and is our
default configuration.

\begin{table}[!h]
\centering
\caption{Cost per step under different chunk sizes $C$.}
\vspace{-2ex}
\label{tab:chunk-eff}
\begin{tabular}{lcccccc}
\toprule
$C$ & 4 & 8 & 16 & 32 & 64 & 128 \\
\midrule
\textbf{Time (s/step)} & OOM & 20.3 & 14.7 & 11.3 & 9.7 & 9.5 \\
\textbf{Memory (GB)}   & OOM & 46.4 & 37.4 & 32.9 & 30.6 & 29.3 \\
\bottomrule
\end{tabular}
\vspace{-1em}
\end{table}

\subsection{Full Tables}
\label{appendix:tables}
\begin{table}[ht]
    \caption{\textbf{Comparison of the Commonsense and General Reasoning Tasks}.
    Comparison of STILL with baseline methods under Llama 3 8B and Llama 3.1 8B models. STILL consistently demonstrates strong performance across benchmarks.
    }
    \label{tab:commonsense-reasoning_full}
    \vspace{-1ex}
    \begin{center}
            \resizebox{1\linewidth}{!}{
                \begin{tabular}{l|c|ccccc|c|cc}
                    \toprule
                    \multirow{2}{*}{\textbf{\textsc{Model}}}
                    & \textbf{\textsc{Training}}
                    & \textbf{\textsc{PIQA}}
                    & \textbf{\textsc{ARC-e}}
                    & \textbf{\textsc{ARC-c}}
                    & \textbf{\textsc{Hella.}}
                    & \textbf{\textsc{Wino.}}
                    & \textbf{\textsc{MMLU}}
                    & \textbf{\textsc{Avg.}}
                    & \textbf{\textsc{Avg.}}
                    \\
                    & \textbf{\textsc{Tokens}} (B)
                    & acc $\uparrow$
                    & acc $\uparrow$ 
                    & acc$_n$ $\uparrow$
                    & acc$_n$ $\uparrow$  
                    & acc $\uparrow$  
                    & (5-shot)  $\uparrow$  
                    & (w. MMLU)
                    & (wo. MMLU)
                    \\
                    \midrule 
                    \rowcolor[HTML]{F1E9E7}
                    \multicolumn{10}{l}{\textbf{\textit{Transformer}}}
                    \\
                    Llama 3 8B~\citep{llama}
                    & 15000$+$
                    & 79.5
                    & 80.0
                    & 53.3
                    & 79.1
                    & 73.1
                    & 65.3
                    & 71.8
                    & 73.0
                    \\
                    Llama 3.1 8B~\citep{llama}
                    & 15000$+$
                    & 81.1
                    & 81.7
                    & 55.1
                    & 79.3
                    & 73.9
                    & 68.0
                    & 74.2
                    & 73.2
                    \\
                    \midrule 
                    \rowcolor[HTML]{F1E9E7}
                    \multicolumn{10}{l}{\textbf{\textit{Subquadratic}}}
                    \\
                    Mamba 8B~\citep{mamba}
                    & 1100
                    & 78.9
                    & 75.4
                    & 42.2
                    & 75.6
                    & 68.3
                    & 28.0
                    & 61.4
                    & 68.1
                    \\
                    Mamba2 8B~\citep{mamba2}
                    & 3500
                    & 79.8
                    & 75.9
                    & 48.1
                    & 77.1
                    & 71.6
                    & 48.7
                    & 67.0
                    & 70.6
                    \\
                    RWKV-6 7B~\citep{rwkv}
                    & 1420
                    & 78.7
                    & 76.8
                    & 46.3
                    & 75.1
                    & 70.0
                    & -
                    & 69.4
                    & 69.4
                    \\
                    TransNormerLLM 7B~\citep{transnormerllm}
                    & 1400
                    & 80.1
                    & 75.4
                    & 44.4
                    & 75.2
                    & 66.1
                    & 43.1
                    & 64.1
                    & 68.2
                    \\
                    Hawk 7B~\citep{griffin}
                    & 300
                    & 80.0
                    & 74.4
                    & 45.9
                    & 77.6
                    & 69.9
                    & 35.0
                    & 63.8
                    & 69.6
                    \\
                    Griffin 7B~\citep{griffin}
                    & 300
                    & 81.0
                    & 75.4
                    & 47.9
                    & 78.6
                    & 72.6
                    & 39.3
                    & 65.8
                    & 71.1
                    \\
                    Falcon3-Mamba 7B~\citep{falcon}
                    & 7300
                    & 80.6
                    & 78.9
                    & 57.1
                    & 80.1
                    & 73.7
                    & 55.1
                    & 70.9
                    & 74.1
                    \\
                    \midrule
                    \rowcolor[HTML]{F1E9E7}
                    \multicolumn{10}{l}{\textbf{\textit{Hybird}}}
                    \\
                    StripedHyena-Nous 7B~\citep{stripedhyena}
                    & -
                    & 78.8
                    & 77.2
                    & 40.0
                    & 76.4
                    & 66.4
                    & 26.0
                    & 60.8
                    & 67.8
                    \\
                    Zamba 7B~\citep{zamba}
                    & 1000
                    & 81.4
                    & 74.5
                    & 46.6
                    & 80.2
                    & 76.4
                    & 57.7
                    & 69.5
                    & 71.8
                    \\
                    RecurrentGemma 9B~\citep{recurrentgemma}
                    & 2000
                    & 80.6
                    & 78.9
                    & 57.1
                    & 80.1
                    & 73.7
                    & 55.1
                    & 70.9
                    & 74.1
                    \\
                    \midrule
                    \rowcolor[HTML]{F1E9E7}
                    \multicolumn{10}{l}{\textbf{\textit{Linearized from Llama 3 8B}}}
                    \\
                    T2R-LoLCATs~\citep{t2r}
                    & 0.04
                    & 62.0
                    & 42.1
                    & 24.7
                    & 32.7
                    & 48.3
                    & 23.2
                    & 38.8
                    & 42.0
                    \\
                    Hedgehog-LoLCATs~\citep{hedgehog}
                    & 0.04
                    & 77.4
                    & 71.1
                    & 40.6
                    & 50.7
                    & 54.3
                    & 24.2
                    & 53.1
                    & 58.8
                    \\
                    SUPRA~\citep{supra}
                    & 20
                    & 78.9
                    & 75.1
                    & 46.5
                    & 71.7
                    & 65.8
                    & 40.9
                    & 63.2
                    & 67.6
                    \\
                    LoLCATs~\citep{lolcats}
                    & 0.04
                    & 80.0
                    & 78.8
                    & 52.2
                    & 75.6
                    & \textbf{73.3}
                    & \underline{43.9}
                    & 67.3
                    & 72.0
                    \\ 
                    Liger-GLA~\citep{liger}
                    & 0.02
                    & \underline{80.3}
                    & \underline{81.1}
                    & \underline{52.5}
                    & \underline{76.3}
                    & 72.0
                    & 43.4
                    & \underline{67.6}
                    & \underline{72.4}
                    \\
                    \rowcolor[HTML]{E2ECF9}
                    STILL~(Ours)
                    & 0.04
                    & \textbf{81.3}
                    & \textbf{81.7}
                    & \textbf{54.8}
                    & \textbf{78.1}
                    & \underline{73.2}
                    & \textbf{59.7}
                    & \textbf{71.5}
                    & \textbf{73.8}
                    \\ 
                    \midrule
                    \rowcolor[HTML]{F1E9E7} 
                    \multicolumn{10}{l}{\textbf{\textit{Linearized from Llama 3.1 8B}}}
                    \\
                    Llamba~\citep{llamba}
                    & 12
                    & \underline{80.9}
                    & 82.5
                    & 54.6
                    & \underline{77.6}
                    & 73.3
                    & \underline{60.0}
                    & \underline{71.5}
                    & \underline{73.8}
                    \\
                    LoLCATs~\citep{lolcats}
                    & 0.04
                    & \underline{80.9}
                    & \underline{82.6}
                    & 54.7
                    & 76.6
                    & \textbf{73.6}
                    & 50.8
                    & 69.9
                    & 73.7
                    \\ 
                    Liger-GLA~\citep{liger}
                    & 0.02
                    & 80.5
                    & 81.8
                    & \underline{55.6}
                    & 75.4
                    & 69.5
                    & 46.9
                    & 68.3
                    & 72.6
                    \\
                    \rowcolor[HTML]{E2ECF9}
                    STILL~(Ours)
                    & 0.04
                    & \textbf{81.3}
                    & \textbf{83.0}
                    & \textbf{56.7}
                    & \textbf{79.0}
                    & \underline{73.4}
                    & \textbf{61.3}
                    & \textbf{72.5}
                    & \textbf{74.7}
                    \\ 
                    \bottomrule
            \end{tabular}
        }
    \end{center}
\end{table}

\begin{table}[!h]
    \centering
    \caption{\textbf{Full Results of the Comparison of the Single Needle-in-a-Haystack (S-NIAH) Tasks}. Accuracy on S-NIAH tasks from the RULER benchmark across context lengths from 0.5K to 4K. Methods are compared under different cache budgets and our approach evaluated under matched or smaller cache constraints.}
    \label{tab:niah_single_full}
    \resizebox{1\linewidth}{!}{
        \begin{tabular}{l|c|llll|llll}
            \toprule
            \multirow{2}{*}{\textbf{\textsc{Model}}}
            & \textbf{\textsc{Cache}}
            & \multicolumn{4}{c|}{\textbf{\textsc{S-NIAH-1}}}
            & \multicolumn{4}{c}{\textbf{\textsc{S-NIAH-2}}}
            \\
            
            & \textbf{\textsc{Tokens}}
            & 0.5K 
            & 1K 
            & 2K 
            & 4K 
            & 0.5K 
            & 1K 
            & 2K 
            & 4K 
            \\
            \midrule 
            SWA & 128 & 0 & 0 & 0 & 0 & 100 & 0 & 0 & 0\\
            LoLCATs~\citep{lolcats} & 128 & 29.4 & 9.6 & 3.0 & 0 & 100 & 17.4 & \textbf{7.2} & \textbf{4.2}\\
            Liger-GLA~\citep{liger} & 128 &  28.4 & 0.2 & 0.2 & 0.2 & 100 & 1.6 & 0.6 & 1.0\\
            \rowcolor[HTML]{E2ECF9}
            STILL~(Ours) & $\leq$128 & \textbf{63.2} & \textbf{22.0} & \textbf{5.8} & \textbf{1.4}& \textbf{100} & \textbf{33.6}& 3.4 & 2.8\\
            \midrule
            SWA & 256 & 66.4 & 4.2 & 0.2 & 0 & 100 & 0 & 0.2 & 0\\
            LoLCATs~\citep{lolcats} & 256 & 84.8 & 26.2 & 10.2 & 2.2 & 100 & 37.0 & 12.2 & 8.2\\
            Liger-GLA~\citep{liger} & 256 & 83.8 & 0.0 &  0.0 & 2.8 & 100 & 1.0 & 4.6 & 0.8\\
            \rowcolor[HTML]{E2ECF9}
            STILL~(Ours) & $\leq$256 & \textbf{94.0} & \textbf{86.6} & \textbf{31.0} & \textbf{10.6} & \textbf{100} & \textbf{63.0} & \textbf{27.2} & \textbf{9.4}\\
            \midrule
            SWA & 512 & 100 & 25.2 & 14.6 & 5.8 & 100 & 44.6 & 0.2 & 0 \\
            LoLCATs~\citep{lolcats} & 512 & 100 & 65.6 & 24.6 & 8.8 & 100 & 71.8 & 17.4 & 16.6\\
            Liger-GLA~\citep{liger} & 512 & 73.4 & 1.0 & 0.0 & 0.0 & 100 & 2.6 & 4.0 & 0.0\\
            \rowcolor[HTML]{E2ECF9}  
            STILL~(Ours) & $\leq$512 & \textbf{100}$_{\textcolor{blue}{+0.0}}$ & \textbf{99.6}$_{\textcolor{blue}{+34.0}}$ & \textbf{89.0}$_{\textcolor{blue}{+64.4}}$ & \textbf{86.2}$_{\textcolor{blue}{+77.4}}$ & \textbf{100}$_{\textcolor{blue}{+0.0}}$ & \textbf{92.6}$_{\textcolor{blue}{+20.8}}$ & \textbf{73.8}$_{\textcolor{blue}{+56.4}}$ & \textbf{37.4}$_{\textcolor{blue}{+20.8}}$\\
            \bottomrule
        \end{tabular}
    }
\end{table}
\begin{table}[ht]
    \centering
    \caption{\textbf{BABILong Benchmark Full Results.}
    Performance comparison on BABILong across increasing context lengths (0K$\sim$4K), evaluating long-context reasoning performance.
    }
    \label{tab:babilong_full}
    \resizebox{1\textwidth}{!}{
        \begin{tabular}{l|cccc|cccc|cccc|cccc|cccc}
            \toprule
            \multirow{2}{*}{\textbf{\textsc{Model}}}
            & \multicolumn{4}{c|}{\textbf{\textsc{QA1}}}
            & \multicolumn{4}{c|}{\textbf{\textsc{QA2}}}
            & \multicolumn{4}{c|}{\textbf{\textsc{QA3}}}
            & \multicolumn{4}{c|}{\textbf{\textsc{QA4}}}
            & \multicolumn{4}{c}{\textbf{\textsc{QA5}}}
            \\
            & 0K 
            & 1K 
            & 2K 
            & 4K 
            & 0K 
            & 1K 
            & 2K 
            & 4K 
            & 0K 
            & 1K 
            & 2K 
            & 4K 
            & 0K 
            & 1K 
            & 2K 
            & 4K 
            & 0K 
            & 1K 
            & 2K 
            & 4K 
            \\
            \midrule 
            SWA & 93 & 2 & 1 & 1 
            & 34 & 3 & 2 & 0
            & 8 & 4 & 2 & 1
            & 74 & 0 & 0 & 0
            & 68 & 18 & 11 & 6
            \\
            LoLCATs & 100 & 22 & 5 & 3 
            & 25 & 10 & 4 & 1 
            & 30 & 17 & 13 & 4 
            & 65 & 21 & 9 & 1 
            & 62 & 42 & 17 & 6 
            \\
            \rowcolor[HTML]{E2ECF9}
            STILL & \textbf{100} & \textbf{45} & \textbf{22} & \textbf{10} 
            & \textbf{57} & \textbf{19} & \textbf{9} & \textbf{5} 
            & \textbf{37} & \textbf{24} & \textbf{15} & \textbf{11} 
            & \textbf{82} & \textbf{59} & \textbf{30} & \textbf{9} 
            & \textbf{83} & \textbf{77} & \textbf{45} & \textbf{24} 
            \\
            \bottomrule
        \end{tabular}
    }
    \vspace{-1em}
\end{table}
\begin{table}[!ht]
\centering
\caption{Extended S-NIAH-1 evaluation up to 8K on Llama 3.1 8B under the
512-token cache budget, STILL retains meaningful retrieval accuracy at 8K, while
all baselines collapse to near zero.}
\label{tab:sniah-8k}
\small
\begin{tabular}{lccccc}
\toprule
S-NIAH-1 & 0.5K & 1K & 2K & 4K & 8K \\
\midrule
SWA         & 100  & 25.2 & 14.6 & 5.8  & 3.0 \\
LoLCATs     & 100  & 65.6 & 24.6 & 8.8  & 0.0 \\
Liger-GLA   & 73.4 & 1.0  & 0.0  & 0.0  & 0.0 \\
STILL (Ours)& 100  & 99.6 & 89.0 & 86.2 & 36.2 \\
\bottomrule
\end{tabular}
\vspace{-1em}
\end{table}

\end{document}